\documentclass[preprint,12pt]{elsarticle}

\usepackage{amssymb}
\usepackage{amsmath}

\usepackage{glossaries} 
\usepackage{tikz} 
\usepackage{relsize} 
\usetikzlibrary{positioning} 
\usepackage{array} 
\usepackage{rotating} 
\usepackage{booktabs} 

\journal{Swarm and Evolutionary Computation}

\newacronym{nlp}{NLP}{Natural Language Processing}
\newacronym{ec}{EC}{Evolutionary Computation}
\newacronym{es}{ES}{Evolutionary Strategies}
\newacronym{ga}{GA}{Genetic Algorithm}
\newacronym{gp}{GP}{Genetic Programming}
\newacronym{dl}{DL}{Deep Learning}
\newacronym{cnn}{CNN}{Convolutional Neural Network}
\newacronym{dnn}{DNN}{Deep Neural Network}
\newacronym{ssl}{SSL}{Self-Supervised Learning}
\newacronym{ne}{NE}{Neuroevolution}
\newacronym{ae}{AE}{autoencoder}
\newacronym{vae}{VAE}{Variational Auto Encoder}
\newacronym{gan}{GAN}{Generative Adversarial Network}
\newacronym{dsge}{DSGE}{Dynamic Structured Grammatical Evolution}
\newacronym{ml}{ML}{Machine Learning}
\newacronym{eml}{EML}{Evolutionary Machine Learning}
\newacronym{essl}{E-SSL}{Evolutionary Self-Supervised Learning}
\newacronym{nas}{NAS}{Neural Architecture Search}
\newacronym{lstm}{LSTM}{Long Short Term Memory}
\newacronym{pso}{PSO}{Particle Swarm Optimisation}
\newacronym{gru}{GRU}{Gated Recurrent Unit}
\newacronym{rnn}{RNN}{Recurrent Neural Network}
\newacronym{brb}{BRB}{Belief Rule Base}
\newacronym{de}{DE}{Differential Evolution}
\newacronym{mse}{MSE}{Mean Squared Error}

\begin{document}

\begin{frontmatter}


\title{Evolutionary Machine Learning meets Self-Supervised Learning: a comprehensive survey}

\author[affiliation]{Adriano Vinhas}
\author[affiliation]{João Correia}
\author[affiliation]{Penousal Machado}
\affiliation[affiliation]{
    organization={University of Coimbra, CISUC/LASI, DEI},
    city={Coimbra},
    country={Portugal}
}

\begin{abstract}
Research that combines Evolutionary Machine Learning and Self-Supervised Learning has been steadily increasing in recent years. Evolutionary Machine Learning has been shown to help automate the design of machine learning algorithms and to lead to more reliable solutions. Self-Supervised Learning, on the other hand, has produced good results in learning useful features when labelled data is limited. This suggests that the combination of these two areas can help both in shaping evolutionary processes and in automating the design of deep neural networks, while also reducing the need for labelled data. However, no detailed surveys exist that explain how Evolutionary Machine Learning and Self-Supervised Learning can be used together. To help with this, we provide an overview of studies that bring these areas together. Driven by the growing interest in this topic and the range of existing works, we consolidate current knowledge on approaches that combine these two fields, propose a definition for this area of research, which we call Evolutionary Self-Supervised Learning, and introduce a taxonomy for it. Finally, we point out some of the main challenges and suggest directions for future research to help Evolutionary Self-Supervised Learning grow and mature as a field.
\end{abstract}



\end{frontmatter}


\section{Introduction}
\label{sec:intro}

\gls{eml}\cite{banzhaf2023handbook, zhang2011evolutionary, telikani2021evolutionary, mirjalili2019evolutionary} is a field that has attracted the attention of researchers due to its ability to produce competitive, and often unexpected, \gls{ml} models that solve a given problem. This ability comes from the fact that \gls{eml} algorithms are stochastic and, therefore, able to promote the emergence of robust solutions under noisy conditions. Additionally, since they don't rely on human-made solutions, they may find unexpected ones. As the field evolves, recent developments have demonstrated competitive or superior performance compared to traditional methods, particularly in terms of generalisation, adaptability, interpretability and efficiency.

In parallel, the use of the \gls{ssl} paradigm has become widespread due to its ability to use unlabelled data to learn representations, which is particularly useful given that the process of data labelling is a task that can be tedious, time-consuming, and prone to errors. Within the text domain, BERT~\cite{kenton2019bert} set new records in several benchmarks by pretraining the model without labels. This success promoted the research in designing pretext tasks with similar impact in computer vision~\cite{tomasev2022pushing}, audio~\cite{chung2021w2v}, video~\cite{wang2023videomae}, or in multimodal scenarios~\cite{hager2023best}.

The growing interest in these fields has sparked a line of research that merges \gls{eml} and \gls{ssl}. This combination can bring benefits by increasing the robustness of the representations learned by \gls{ssl} algorithms or improving the evolution process behind the learned solutions, especially in situations where labelled data is limited.

\begin{figure}[!ht]
    \centering
    \includegraphics[width=1.\linewidth]{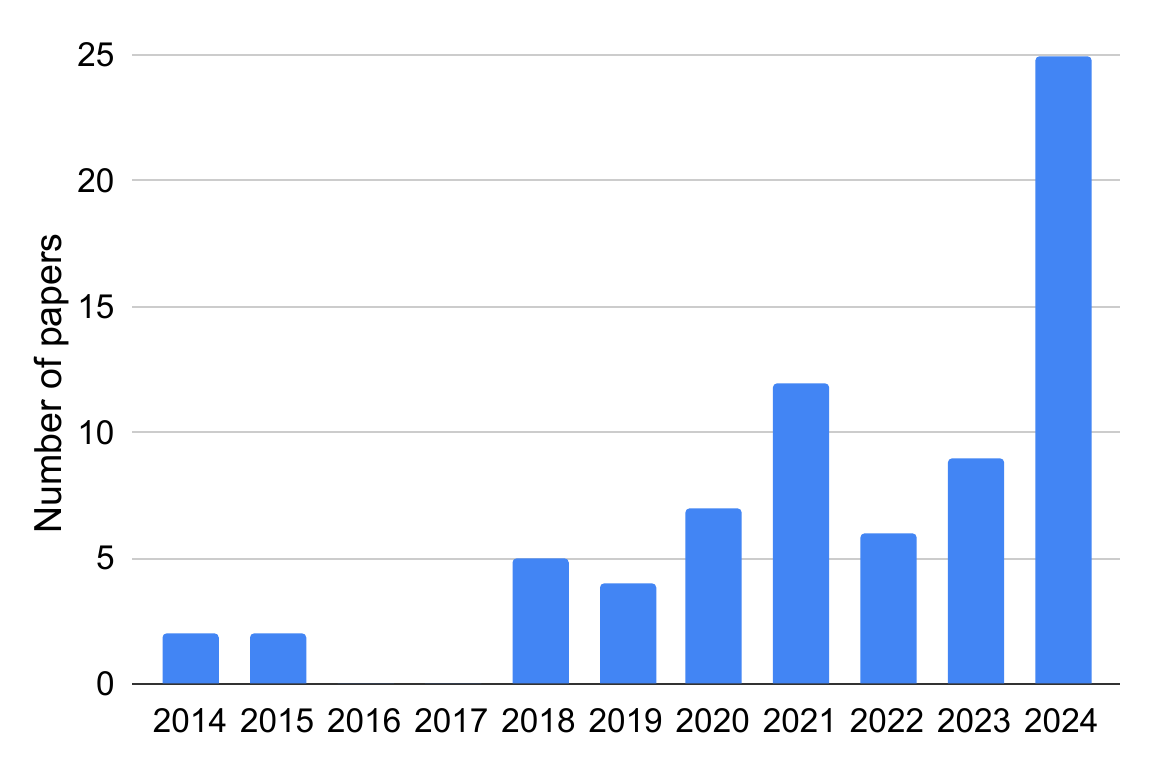}
    \caption{Number of surveyed publications related with \gls{essl} breakdown by year}
    \label{fig:surveyed_papers}
\end{figure}

The surveys about \gls{eml} that were identified~\cite{zhang2011evolutionary, telikani2021evolutionary, mirjalili2019evolutionary} do not handle the topic of leveraging unlabelled data for representation learning despite listing a few studies that fit the subject. Similarly, surveys on \gls{ssl}~\cite{technologies9010002, gui2024survey} fall short in terms of detailing the efforts to automate the search or design of pretext tasks. Interestingly, conda activate neat covers the intersection of Supervised Learning with \gls{eml}~\cite{miikkulainen2023evolutionary}. However, to the best of our knowledge, no survey focuses on works that intersect \gls{eml} and \gls{ssl} with such a broad view. Nevertheless, we have identified a survey that focuses specifically on the use of \gls{ec} in the context of \glspl{gan}~\cite{wang2024application}. Since a dedicated survey already covers this topic, we exclude it from our scope and direct readers to it for further details.
In the scope of this paper, we focus on studies that combine \gls{eml} and \gls{ssl}, defining a field which encompasses all the works that fit within this intersection, naming it \gls{essl}. In total, we identified 72 papers that fall within the scope of this work. The number of these publications by year is shown in Figure~\ref{fig:surveyed_papers}, and it supports the idea that interest in this area is growing.
To find these papers, we used search terms related to \gls{ec} together with terms related to \gls{ssl}, using the AND operator. Some of the \gls{ec} terms include: evolutionary, evolutionary neural architecture search, genetic programming, evolutionary strategies, and cma-es. For the \gls{ssl} keywords, we used terms like self-supervised, pretext task, autoencoder, contrastive, siamese networks, and unsupervised pretraining. We ran our searches on Google Scholar, IEEExplore, ACM Digital Library, and SpringerLink.

The main contributions of this survey are: (i) the formalisation of Evolutionary Self-Supervised Learning (E-SSL) as a research area, together with a taxonomy derived from recent studies; and 
(ii) a discussion of open challenges and promising directions for future work within this taxonomy.

The remainder of this paper is organised as follows: Section~\ref{sec:ssl} gives background on the \gls{ssl} field, while Section~\ref{sec:eml} provides an overview of the \gls{eml} field; Section~\ref{sec:essl} describes the new research area and the proposed taxonomy; Section~\ref{sec:discussion_future_research} covers open challenges and future research directions; Finally, Section~\ref{sec:conclusion} wraps up the paper by summarising the main contributions.

\section{Self-Supervised Learning (SSL)}
\label{sec:ssl}
\gls{ssl} has been gaining popularity in recent years due to its ability to learn from large amounts of unlabelled data, which is often easier to obtain than labelled data. It is a learning paradigm that uses the input data itself to create pseudo-labels that will guide the learning process. Pseudo-labels can be defined based on a property of the input, or generated from the relationship between the original input and a modified version of it. This modification can take the form of parts of the original input, a corrupted version (created by masking, adding noise, or applying some other distortion), or even a different modality of the data. The goal of~\gls{ssl} is to learn representations from signals derived from the inputs~\cite{de1994learning}.

\begin{figure}[!ht]
    \centering
    \includegraphics[width=0.95\linewidth]{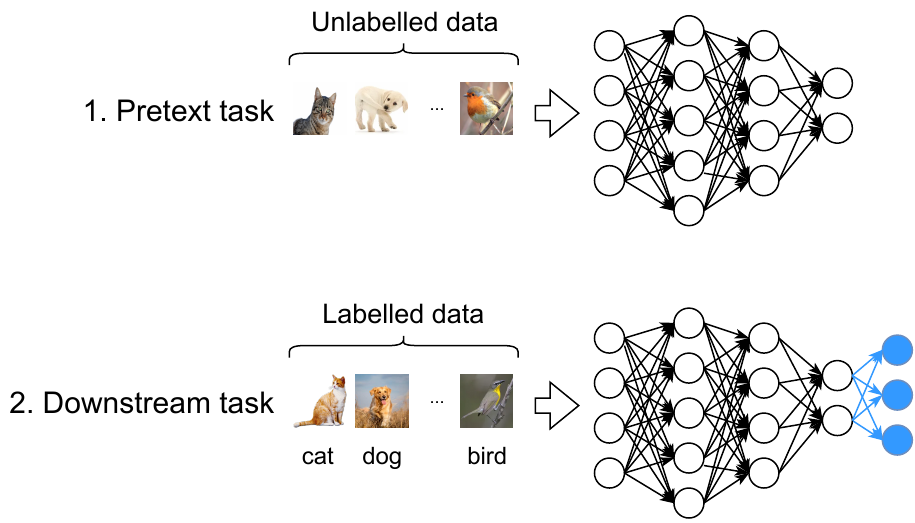}
    \caption{Overview of the \gls{ssl} process for an image classification problem.}
    \label{fig:ssl_process}
\end{figure}

The \gls{ssl} paradigm is particularly useful when the number of labelled inputs is limited, as it can be divided into two distinct stages. In the first stage, a model is trained without using any labels and without explicitly considering the final goal. Instead, pseudo-labels are used to supervise training, which forces the model to learn how to extract features for a task that is related to the actual goal but not the same. This initial step is known as the pretext task. The second stage, called the downstream task, uses the model trained in the pretext task to help solve the actual target task. This stage follows a Supervised Learning setting, where features extracted from labelled inputs are used to train the model that solves the target task. The representations learned during the pretext task can either be reused directly or fine-tuned during this second step. An overview of this two-step process in \gls{ssl} is shown in Figure~\ref{fig:ssl_process}.

The success of \gls{ssl} depends on the quality of the representations learned during the pretext task. According to Bengio et al.~\cite{bengio2013representation}, good representations have three main characteristics. First, they should be \textbf{expressive}, meaning they can cover a large portion of the input space while using a lower-dimensional representation. Second, they should be \textbf{disentangled}, meaning each factor of variation in the input is clearly separated. For example, an image may contain variation factors such as object identity, position, lighting, and pose. Finally, good representations should be \textbf{invariant}, meaning they remain stable even when transformations are applied to the input.

One critical aspect of \gls{ssl} is the design of the pretext task, as it affects the quality of the learned representations. One of the main challenges in designing a pretext task is making sure that the network does not produce the same representation for several different inputs, an issue known as collapse~\cite{li2022understanding}. For example, this phenomenon can occur if the loss function is defined based only on the similarity to other inputs, as the network would learn a constant representation that would minimise the distance to them.

Most early forms of \gls{ssl} design pretext tasks by manipulating datasets to create a proxy dataset. The pretext task then follows a Supervised Learning setting using the artificially generated labels for the proxy data. Doersch et al.~\cite{doersch2015unsupervised} divide each image into a 3×3 grid and extract pairs of patches -- one from the center and the other from one of the eight surrounding positions. Each pair is labelled with one of eight possible classes. Gidaris et al.~\cite{komodakis2018unsupervised} rotate images by 0°, 90°, 180°, or 270°, and train a neural network to predict the applied rotation. Larsson et al.~\cite{larsson2016learning} provide equalised grayscale images to a neural network, which is trained to predict hue and chroma distributions for each pixel, turning the pretext task into a colourisation problem.
Dosovitskiy et al.~\cite{7312476} take a contrastive approach by formulating a multi-class classification task. Each original instance, along with several modified versions, is treated as a single class. In contrast to these classification-based approaches, Li et al.~\cite{li2021generic} design a regression task in which feature maps are compared to multiple synthetic signals. Zhang et al.~\cite{zhang2021neural} assign random labels to instances, either by shuffling existing labels or generating new ones from a discrete uniform distribution.

Within the video domain, the temporal dimension introduces new ways to craft pretext tasks. For instance, Pickup et al.~\cite{pickup2014seeing} propose reversing the order of video frames to create a binary classification task, where videos are classified as playing forward or backward. This idea was later used as a pretext task by Piergiovanni et al.~\cite{piergiovanni2020evolving}. Misra et al.~\cite{misra2016shuffle} sample tuples of video frames that may be either in the correct order or shuffled, crafting a binary classification task based on temporal order.

The audio modality can also be used to train representations. Arandjelovic and Zisserman~\cite{arandjelovic2017look} train a network to detect whether a given audio clip matches the corresponding video or not. Similarly, Korbar et al.~\cite{korbar2018cooperative} feed pairs of audio and visual data into a model and train it to detect whether the audio is synchronised with the video.

Another way to craft pretext tasks is by using the original dataset but modifying the labels through unsupervised learning during training. DeepCluster~\cite{caron2018deep} trains representations by applying K-Means clustering to the learned features and then using the cluster assignments as pseudo-labels to update network weights. Yang et al.~\cite{yang2016joint} follow a similar approach, but use agglomerative clustering instead. Zhang et al.~\cite{zhuang2019local} use clustering not to assign labels directly, but to define local neighbourhoods. For each representation z, the model is trained to adjust the network weights so that more of the points in the same neighbourhood also appear in the set of k nearest neighbours of z.

Generative architectures are also an option for crafting pretext tasks. For instance, \glspl{ae} are architectures trained through information restoration. The goal is to encode the input into a compressed representation and then reconstruct the original input by decoding that representation. In this case, the pseudo-label is the input itself. However, the robustness of the learned representations can be improved by feeding deteriorated versions of the input to the encoder~\cite{vincent2008extracting}. An example of this is the context \gls{ae}~\cite{pathak2016context}, which trains on images with large missing regions, forcing the model to learn image in-painting.

Zhang et al.~\cite{zhang2017split} propose an \gls{ae} split into two parts: one receives grayscale images and predicts their colour version, while the other learns the reverse process. He et al.~\cite{he2022masked} propose masking image patches and applying the encoder only to the visible parts, while the decoder reconstructs the input using both the visible and masked regions. In the video domain, Srivastava et al.~\cite{srivastava2015unsupervised} train an \gls{ae} where both the encoder and decoder are \glspl{lstm}. The goal of the pretext task is to predict the next video frame based on a sequence of previous frames.

Alternatively, pretext training can be done with \glspl{gan}. In this architecture, representations from the generator are learned by training a discriminator to distinguish between real and generated (fake) images. In some cases, the generator is also trained to solve an additional task, such as predicting image rotation~\cite{chen2019self}.

Many recent state-of-the-art methods rely on higher-level architectures with multiple branch networks, most commonly dual-branch networks. These branches can be either asymmetric or siamese, sharing the same architecture or even the same weights. In simple terms, different views or parts of the same input are given to each branch to produce invariant representations. In this context, Noroozi and Favaro~\cite{noroozi2016unsupervised} proposed teaching a network to solve jigsaw puzzles to help it learn useful features. To do this, images are divided into nine tiles, which are randomly shuffled and passed through a siamese-ennead network trained to predict the correct arrangement.

Another common approach to learning representations is contrastive learning. In this setup, the goal is to bring representations of similar inputs (positives) closer together while pushing those of different inputs (negatives) further apart. PIRL~\cite{misra2020self} uses a memory bank to compute the loss: for each image $i$, it encourages the representations from two branches to match the stored vector $m(i)$, while forcing them to be as dissimilar as possible to a set of sampled negatives $m_{j\neq i}$.
SimCLR~\cite{chen2020simple} provides two views of the same input to a siamese network that was trained using the same contrastive principle. It was shown that a memory bank is not required if the batch size is large enough. NNCLR~\cite{dwibedi2021little} executes a slightly different version whereby one of the branches does not use the respective input representation to compute loss. Instead, it computes the nearest neighbours of that input representation, in order to provide more semantic variations to sample the positives.

As an alternative to contrastive approaches, representations can be trained using correlation-based metrics. For example, Barlow Twins~\cite{zbontar2021barlow} uses the same siamese network structure, but the loss function is based on the cross-correlation matrix between the representations of the two branches. The goal is to maximise the invariance of features across different views of the same input, while reducing redundancy between features. This can be seen as a contrastive objective applied at the feature level rather than the instance level.
VicReg~\cite{bardes2021vicreg} extends Barlow Twins by adding an extra objective to maintain sufficient variance in each feature across a batch. It also promotes invariance by computing the \gls{mse} between the two views of the same image, instead of relying on the diagonal of the cross-correlation matrix.

Self-distillation is another approach to representation learning using dual-branch networks. In this context, one branch acts as the teacher, while the other is the student. The student uses the output from the teacher as a pseudo-label and learns online through gradient descent. The teacher network learns by transferring some of the knowledge from the student network, a process known as distillation. In practice, weight updates on the teacher's side are performed based on a moving average of the student network weights. 

BYOL~\cite{grill2020bootstrap} uses two asymmetric branches: the encoders do not share weights, and the student branch includes an additional predictor that aims to match the output of the teacher. SimSiam~\cite{chen2021exploring} showed that representation collapse can be avoided even when the encoders share weights, as long as gradient flow is stopped from reaching the teacher branch.
IBOT~\cite{zhou2021ibot} incorporates masked image modelling into the student branch by training it to predict masked tokens from unmasked ones. I-JEPA~\cite{assran2023self} adopts an asymmetric design to train a student network using the context from an image. Based on this context, the student predicts separate representations for different image regions, which are then compared to the target representations provided by the teacher. This method was later adapted to the video domain~\cite{bardes2024revisiting}.

\section{Evolutionary Machine Learning (EML)}
\label{sec:eml}

\gls{eml} is a field that uses the synergies between \gls{ec} and \gls{ml} to enhance aspects of each other. The adoption of \gls{eml} approaches has grown substantially over the last years due to their extra flexibility and motivation to automate the design of \gls{ml} models. \gls{eml} has a wide range of applications, including environmental science, medicine, finance, and robotics.

Bhanzaf et al.~\cite{banzhaf2023handbook} acknowledges that each of the areas benefits the other when combined, hence providing a broader view over the field. \gls{eml} works can be categorised based on three types of interaction between \gls{ec} and \gls{ml}. The first category encompasses works that apply \gls{ec} for \gls{ml} methods. The second category comprises works that apply \gls{ec} as an ML method. Within this category, \gls{ec} is used directly as an \gls{ml} model. Finally, the third category covers a range of studies where \gls{ml} can be used within \gls{ec} algorithms.

In the first two categories, \gls{ec} is brought into \gls{ml}, carrying several benefits which explain the superior performance of \gls{eml} compared to traditional \gls{ml}. \gls{ec} brings an element of surprise which comes from its stochastic nature. Since \gls{ec} is driven by the survival of the fittest principle, this stochastic nature will help (i) exploiting flaws in a system, and (ii) enhancing the robustness of the final solution, which emerged from noisy starting points. 
Additionally, an \gls{ec} algorithm can be used to optimise towards several goals simultaneously, meaning that one can evolve an \gls{ml} model towards the best accuracy possible, but also taking into account other metrics such as training speed, inference speed and number of parameters.


As for the third category, it brings \gls{ml} into \gls{ec}. One possible role of \gls{ml} in \gls{ec} is to improve specific parts of the evolutionary process -- such as genotype-to-phenotype mapping, fitness evaluation, variation operators (e.g. mutation and crossover), parent selection, or replacement strategy.



For instance, the fitness function can be replaced by an \gls{ml} model that assigns the fitness of an individual~\cite{correia2016xfaces}, use LLMs to perform ``intelligent'' crossover~\cite{romera2024mathematical}, or learn new genotype-phenotype mappings~\cite{tian2020solving}. 
Additionally, components from \gls{ec} algorithms come with parameters, such as the probability of applying crossover or mutation operators, and the population size. Since these parameters affect the evolutionary process, \gls{ml} can be used to select suitable values -- either during evolution or \textit{a priori}. \gls{ml} can also be applied to the outputs of an \gls{ec} algorithm to summarise, filter, or select results.

\section{Evolutionary Self-Supervised Learning (E-SSL)}
\label{sec:essl}

\gls{essl} can be seen as a subset of \gls{eml}. Concretely, it concerns the application of \gls{ec} to \gls{ml} models that learn through \gls{ssl}. The stochastic element brought by \gls{ec} is a factor that can enhance the quality of the representations learned during the pretext task, as well as exploit any of their potential shortcomings. Moreover, \gls{ec} is able to do the same for the downstream task.

\begin{figure}[!ht]
    \centering
    \includegraphics[width=.7\linewidth]{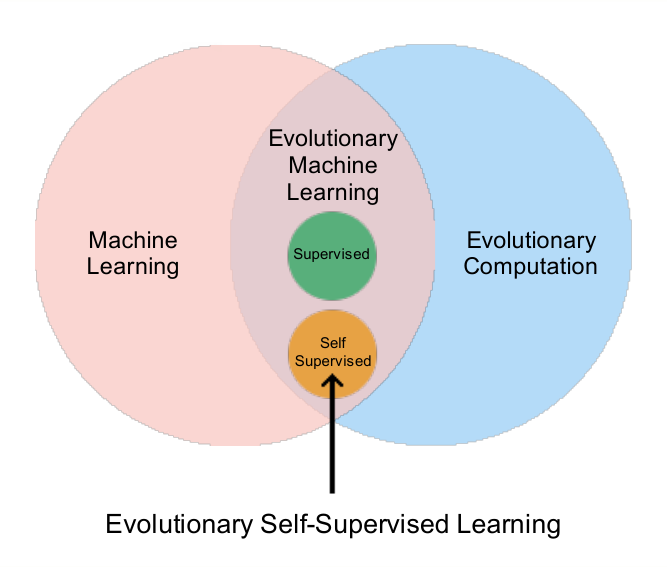}
    \caption{Relation of \gls{essl} field when compared to \gls{ml}, \gls{ec} and \gls{eml}}
    \label{fig:essl_structure}
\end{figure}

In order to contextualise \gls{essl} and relate it with other fields, a visual diagram that explains how \gls{essl} is positioned against related areas like \gls{ml}, \gls{ec} and \gls{eml} is depicted in Figure~\ref{fig:essl_structure}. In this figure, one can see the conceptual overlaps and distinctions among these fields.

Since we consider \gls{essl} to be a subset of \gls{eml}, some aspects of the \gls{eml} taxonomy proposed by Banzhaf et al.~\cite{banzhaf2023handbook} can be reused in this context. However, since \gls{ssl} is typically divided into two stages, this opens the possibility of introducing evolution at different points in the learning process. Based on this, \gls{essl} works are grouped into two main categories: those that use \gls{ec} to support \gls{ssl} (Section~\ref{subsec:ec_for_ssl}), and those that apply \gls{ssl} within \gls{ec} algorithms (Section~\ref{subsec:ssl_for_ec}).

\subsection{EC for SSL}
\label{subsec:ec_for_ssl}

The application of \gls{ec} algorithms to \gls{ssl} can be viewed along two distinct dimensions. The first concerns the stage of the learning process that \gls{ec} targets. As noted in Section~\ref{sec:ssl}, \gls{ssl} is typically divided into two tasks -- pretext and downstream -- and \gls{ec} can be applied to either. The second dimension relates to the specific component of the \gls{ssl} algorithm being optimised.

Regardless of the stage being targeted, three main components of the optimisation process can be identified: the dataset used to train the model; the topology, which refers to the structural aspects of the model; and learning, which defines the algorithm responsible for optimising the model’s parameters. \gls{ec} can be treated as a black-box optimiser that operates on one or more of these components at any stage of the learning process.

In this section, works are grouped according to the \gls{ssl} task where \gls{ec} is applied. Since \gls{ec} is more frequently used in the context of pretext tasks, studies targeting pretext are further organised based on the components mentioned above -- dataset, topology, and learning. In addition, some works that address multiple components at once are also identified.

The works that fall under this section are described in Table~\ref{tab:works_ec_for_ssl}. These studies are characterised in terms of components targeted by \gls{ec}, and how the evolution engages with \gls{ssl}. More specifically, we identify the type of \gls{ssl} goals defined for the problem, whether the \gls{ssl} objective type is directly customised by the evolution process, the stage at which \gls{ec} is applied. Additionally, we focus on identified how the evolution process evaluated the candidate solution, as well as highlighting whether the work developed by the authors conducted test to assess representation robustness in scenarios of label scarcity. 

\renewcommand{\arraystretch}{1}
\begin{sidewaystable}
\scriptsize
\centering
\begin{tabular}{>{\centering\arraybackslash}m{2.2cm}>{\centering\arraybackslash}m{3.3cm}>{\centering\arraybackslash}m{3cm}>{\centering\arraybackslash}m{1.8cm}>{\centering\arraybackslash}m{2.6cm}>{\centering\arraybackslash}m{1.6cm}>
{\centering\arraybackslash}m{2.7cm}>{\centering\arraybackslash}m{1cm}}\toprule
 & & \multicolumn{4}{c}{\textbf{EC engagement with SSL}} & \multicolumn{2}{c}{\textbf{Evaluation}} \\\cmidrule(r){3-6}\cmidrule(lr){7-8}
\textbf{Work name} & \textbf{What is evolved} & \textbf{SSL goal type} & \textbf{SSL customised by EC} & \textbf{How} & \textbf{When is EC applied} & \textbf{Fitness metric} & \textbf{Label scarcity study} \\
\hline
CGP-NASV2 \cite{garcia2024progressive} & Encoder & Rotation prediction & No & - & During pretext & Pretext accuracy & No \\
\hline
EvoDeNSS \cite{vinhas2024towards} & Encoder, Learning hyperparams & Feature correlation & No & - & During pretext & Downstream accuracy & Yes \\
\hline
ELo \cite{piergiovanni2020evolving} & Loss function & Multiple (combined) & Yes & Linear combination of SSL losses & During pretext & Zipf law fit metric & Yes \\
\hline
Barrett et al. \cite{barrett2023evolutionary} & Data augmentation policy & Multiple (separately) & No & - & During downstream & Downstream accuracy & No \\
\hline
CSENet \cite{liu2024collaborative} & Loss function & Multiple (combined) & Yes & Linear combination of SSL losses & During pretext & Downstream accuracy & No \\
\hline
Li et al. \cite{li2022does} & Loss function, Data augmentation policy & Multiple (separately) & No & - & During pretext & Agent Score & No \\
\hline
FaUNAE \cite{xue2021fast} & Encoder & Contrastive & No & - & During pretext & Validation loss (pretext) & No \\
\hline
AutoSSL \cite{jin2021automated} & Loss function & Multiple (combined) & Yes & Linear combination of SSL losses & During pretext & Homophily based on cluster assignments & No \\
\hline
AutoNAS \cite{han2022autonas} & Encoder & Input reconstruction & No & - & During pretext & Validation loss (pretext) & No \\
\hline
Shen et al. \cite{shen2024evolutionary} & Learning hyperparams & Rotation prediction + Color arrangement prediction (combined) & No & - & During downstream & Not reported & No \\
\hline
GenNAS \cite{li2021generic} & Data labels & Feature maps prediction & Yes & Generation of pseudo-labels & During pretext & Validation loss (Pretext) & No \\
\hline
GenNAS with Eproxy \cite{li2023extensible} & Data labels, Learning hyperparams, Projector & Feature maps prediction & Yes & Generation of pseudo-labels & During pretext & Validation loss (Pretext) & No \\
\hline
RLNAS \cite{zhang2021neural} & Encoder & Random label prediction & No & - & During pretext & Convergence based metric & No \\
\hline
StGP \cite{rodrigues2023performance} & Downstream network & Multiple (separately) & No & - & During downstream & F1 score & No \\
\hline
David and Greental \cite{david2014genetic} & Encoder & Input reconstruction & No & - & During pretext & Train loss (Pretext) & No \\
\hline
EvoAE \cite{lander2015evoae} & Encoder & Input reconstruction & No & - & During pretext & Validation loss (Pretext) & No \\
\hline
Assunção et al. \cite{assuncao2018automatic} & Encoder, Projector & Input reconstruction & No & - & During pretext & Downstream accuracy + Net structure metric & No \\
\hline
EvoVAE \cite{chen2020evolving} & Encoder, Projector & Input reconstruction & No & - & During pretext & Train loss (pretext) & Yes \\
\hline
EUDNN \cite{sun2018evolving} & Encoder, Projector & Evolved & Yes & Weights evolved & During pretext & Downstream accuracy & No \\
\hline
SEvoAE \cite{cai2018novel} & Encoder & Input reconstruction & No & - & During pretext & L1 validation loss + L2 validation loss (Pretext) & No \\
\hline
\end{tabular}
\caption{Works where \gls{ec} is applied for \gls{ssl}}
\label{tab:works_ec_for_ssl}
\end{sidewaystable}
\renewcommand{\arraystretch}{1}

\renewcommand{\arraystretch}{1}
\begin{sidewaystable}
\scriptsize
\centering
\begin{tabular}{>{\centering\arraybackslash}m{2.2cm}>{\centering\arraybackslash}m{3.3cm}>{\centering\arraybackslash}m{3cm}>{\centering\arraybackslash}m{1.8cm}>{\centering\arraybackslash}m{2.6cm}>{\centering\arraybackslash}m{1.6cm}>
{\centering\arraybackslash}m{2.7cm}>{\centering\arraybackslash}m{1cm}}\toprule
     & & \multicolumn{4}{c}{\textbf{EC engagement with SSL}} & \multicolumn{2}{c}{\textbf{Evaluation}} \\\cmidrule(r){3-6}\cmidrule(lr){7-8}
\textbf{Work name} & \textbf{What is evolved} & \textbf{SSL goal type} & \textbf{SSL customised by EC} & \textbf{How} & \textbf{When is EC applied} & \textbf{Fitness metric} & \textbf{Label scarcity study} \\
\hline
Ludwig and Claes \cite{ludwig2023compressing} & Encoder & Masking + Contrastive & No & - & During downstream & Word Error Rate & No \\
\hline
MaskTAS \cite{yan2024masked} & Encoder & Masking + Distillation & No & - & During pretext & Feature similarity between student and teacher & No \\
\hline
Kanwal et al. \cite{kanwal2021evolving} & Encoder & Input reconstruction & No & - & During pretext & Validation loss (pretext) + Downstream accuracy + Net structure & No \\
\hline
ST-ITO \cite{steinmetz2024st} & Downstream coefficients & Audio effect and preset prediction & No & - & During downstream & Cosine similarity & No \\
\hline
Sors et al. \cite{sors2021simple} & Loss function & Contrastive & Yes & Linear combination of loss components & During pretext & Downstream mean average precision & No \\
\hline
LTR-CE \cite{sors2021simple} & Data sample pairs & Masking & No & - & Before downstream & Validation loss (downstream) & No \\
\hline
CLAMP-ViT \cite{ramachandran2024clamp} & Encoder & Contrastive & No & - & After pretext & Validation loss (pretext and downstream) & No \\
\hline
Evol-Q \cite{frumkin2023jumping} & Encoder & Contrastive & No & - & After pretext & Validation loss (downstream) & No \\
\hline
LPQ \cite{ramachandran2024algorithm} & Encoder & Contrastive & No & - & After pretext & Validation loss (pretext) + Net structure & No \\
\hline
JCLB \cite{hu2024joint} & Downstream network & Contrastive + Masking & No & - & During downstream & Downstream accuracy & No \\
\hline
AOC-VAE \cite{shang2024efficient} & Encoder, Projector & Input reconstruction & No & - & During pretext & Validation loss (pretext) & Yes \\
\hline
ADWSL \cite{cheng2020adaptive} & Learning hyperparams & Input reconstruction & No & - & During pretext & Train loss (pretext) & Yes \\
\hline
FCAE \cite{sun2018particle} & Encoder, Projector & Input reconstruction & No & - & During pretext & Train loss (pretext) & Yes \\
\hline
MONCAE \cite{dimanov2021moncae} & Encoder, Projector & Input reconstruction & No & - & During pretext & Validation loss (pretext) & No \\
\hline
Hajewski and Oliveira \cite{hajewski2020evolving, hajewski2021distributed} & Encoder, Projector & Input reconstruction & No & - & During pretext & Validation loss (pretext) & No \\
\hline
Charte et al. \cite{charte2019automating, charte2020evoaaa} & Encoder, Projector, Loss function & Input reconstruction & No & - & During pretext & Train loss (pretext) & No \\
\hline
eVAE \cite{wu2024evae} & Loss function & Input reconstruction & No & - & During pretext & Train loss (pretext) & No \\
\hline
MP\textsuperscript{2} \cite{sun2023multitask} & Prompts & unknown & No & - & During downstream & unknown & No \\
\hline
SA-eInfoVAE \cite{emm2024self} & Loss function & Input reconstruction & No & - & During pretext & Train loss (pretext) & No \\
\hline
\end{tabular}
\end{sidewaystable}
\renewcommand{\arraystretch}{1}

\renewcommand{\arraystretch}{1}
\begin{sidewaystable}
\scriptsize
\centering
\begin{tabular}{>{\centering\arraybackslash}m{2.2cm}>{\centering\arraybackslash}m{3.3cm}>{\centering\arraybackslash}m{3cm}>{\centering\arraybackslash}m{1.8cm}>{\centering\arraybackslash}m{2.6cm}>{\centering\arraybackslash}m{1.6cm}>
{\centering\arraybackslash}m{2.7cm}>{\centering\arraybackslash}m{1cm}}\toprule
     & & \multicolumn{4}{c}{\textbf{EC engagement with SSL}} & \multicolumn{2}{c}{\textbf{Evaluation}} \\\cmidrule(r){3-6}\cmidrule(lr){7-8}
\textbf{Work name} & \textbf{What is evolved} & \textbf{SSL goal type} & \textbf{SSL customised by EC} & \textbf{How} & \textbf{When is EC applied} & \textbf{Fitness metric} & \textbf{Label scarcity study} \\
\hline
Tabak et al. \cite{tabak2024evolutionary} & Encoder & Input reconstruction & No & - & During pretext & Loss (Pretext) & No \\
\hline
Preen et al. \cite{preen2021autoencoding} & Encoder, Learning hyperparams & Input reconstruction & No & - & During pretext & Downstream accuracy & No \\
\hline
Bu et al. \cite{bu2024layer} & Learning hyperparams & Masking & No & - & During downstream & Predicted validation accuracy from surrogate model & No \\
\hline
Hajewski et al. \cite{hajewski2020evolutionary, hajewski2021efficient} & Encoder & Input reconstruction & No & - & During pretext & Validation loss (pretext) & No \\
\hline
GP autoencoder \cite{rodriguez2019evolving} & Encoder, Projector & Input reconstruction & No & - & During pretext & Loss (pretext) & No \\
\hline
GPE-AE \cite{schofield2023genetic} & Encoder & Input reconstruction & No & - & During pretext & Loss (pretext) & No \\
\hline
EvoMAE \cite{ha2021evolving} & Encoder, Downstream network & Input reconstruction & No & - & During pretext + During downstream & Validation accuracy (Downstream) & No \\
\hline
E-CAE \cite{suganuma2018exploiting} & Encoder, Projector & Input reconstruction & No & - & During pretext & Peak Signal
to Noise ratio on validation set (pretext) & No \\
\hline
SAENE \cite{silhan2019evolution} & Encoder, Projector & Input reconstruction & No & - & During pretext & Local Continuity Meta-Criterion metric on validation set (pretext) & No \\
\hline
\end{tabular}
\end{sidewaystable}
\renewcommand{\arraystretch}{1}

\subsubsection{Pretext task -- Dataset}

A limited number of studies focus on evolving components that affect the dataset used in the pretext task. Among the works identified, two main groups can be distinguished: one where the \gls{ec} algorithm optimises the pseudo-labels assigned to the data, and another where evolution targets the inputs. In the latter case, the evolutionary process can operate directly on the inputs -- through generation or combinatorial optimisation -- or act on a function $f$ that modifies the original data.

\begin{figure}[!ht]
    \centering
    \includegraphics[width=1.\linewidth]{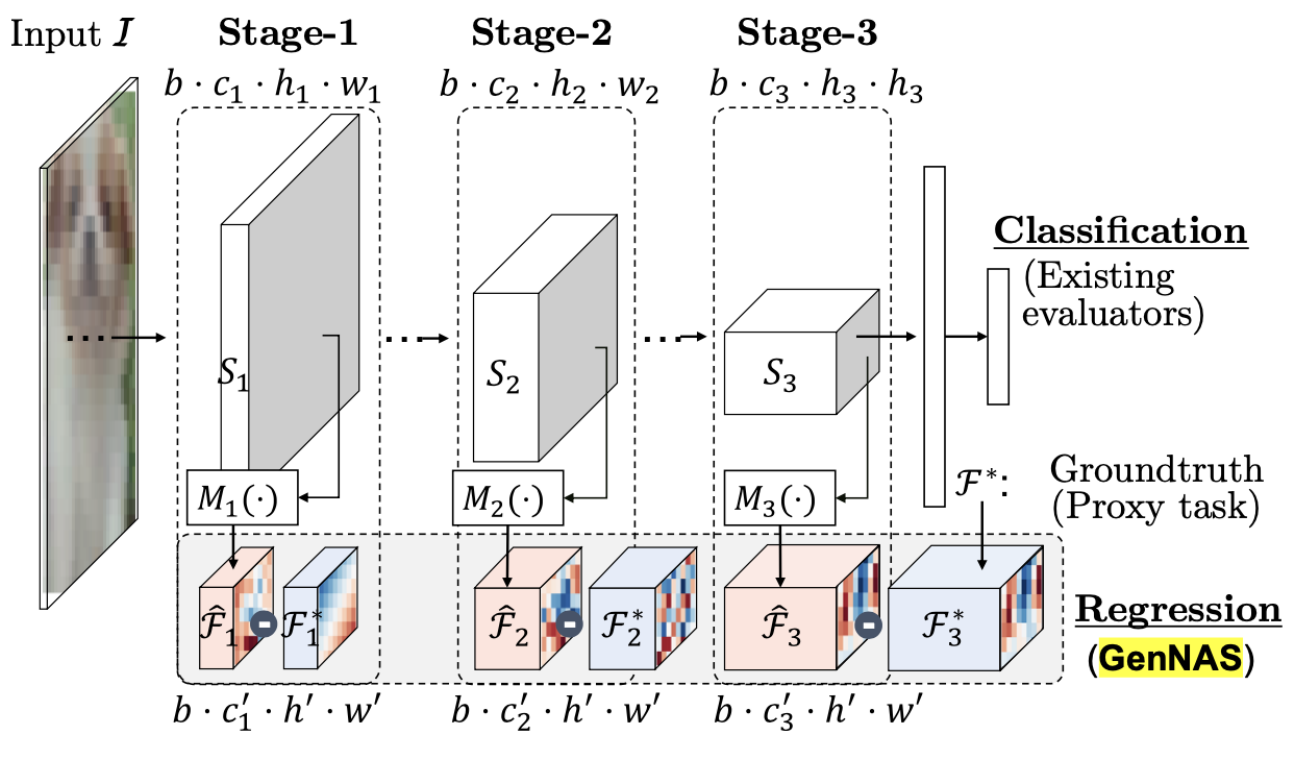}
    \caption{GenNAS for convolutional neural network architectures~\cite{li2021generic}. Each stage contains a single convolutional layer $M$ whose feature maps are optimised to be as approximated as possible to synthetic signals.}
    \label{fig:gennas}
\end{figure}

In the first group, Li et al.~\cite{li2021generic} apply a \gls{ec} algorithm to explore which combinations of pseudo-labels are most effective for representation learning in the GenNAS framework. GenNAS trains a network to learn representations by having its feature maps approximate synthetic signals, as illustrated in Figure\ref{fig:gennas}. A convolutional layer is added to each stage of the network, and a loss function is designed to match the output of each auxiliary layer to a synthetic signal. The authors use a \gls{ga} to evolve parameters that generate the synthetic signals at each stage, evaluating the quality of the learned representations on a small set of network architectures. The assumption is that the closer the network’s output is to the evolved target signal at each stage, the better the solution.

The best set of synthetic signals is then transferred to several fixed architectures and to multiple cell-based \gls{nas} search spaces. In the latter case, the outcome is a set of evolved cells used to build a final architecture following a predefined skeleton, forcing another training round on the whole structure.

Barrett et al.~\cite{barrett2023evolutionary} evolve data augmentation components in the image domain. They assume that, given a set of available augmentation functions, the intensity values of these functions can be optimised during the pretext task. The authors encode both the augmentation functions and their intensity levels, and use an \gls{ec} algorithm to find the optimal data augmentation configuration based on performance in the downstream task.

The selected augmentation functions and their corresponding intensities are tested separately across different \gls{ssl} algorithms and compared to a baseline. In addition, the authors explore a more holistic approach, where the \gls{ssl} algorithm itself is encoded into the genotype of each individual, with optimisation again guided by accuracy on the downstream task.

\subsubsection{Pretext task -- Topology}

A significant portion of \gls{essl} studies focus on optimising model topology using \gls{ec}, as the structure of the model plays a key role in task performance. One of the most important factors when evolving topologies is the size of the search space, as it affects how an \gls{ec} algorithm converges toward optimal regions. In this context, Tabak et al.~\cite{tabak2024evolutionary} constrain the evolution of \glspl{ae} in several ways. They use a \gls{ga} with a variable-length evolutionary representation, where each gene encodes the number of neurons in a layer (restricted to dense layers only).

To reduce the search space, the decoder architecture is fixed in advance, meaning that only the encoder is evolved. The \glspl{ae} are evolved to design Item Response Theory (IRT) models, with reconstruction loss as the guiding objective. These models are intended to estimate traits or abilities of students based on their responses to questions.

\begin{figure}[!ht]
    \centering
    \includegraphics[width=1.\linewidth]{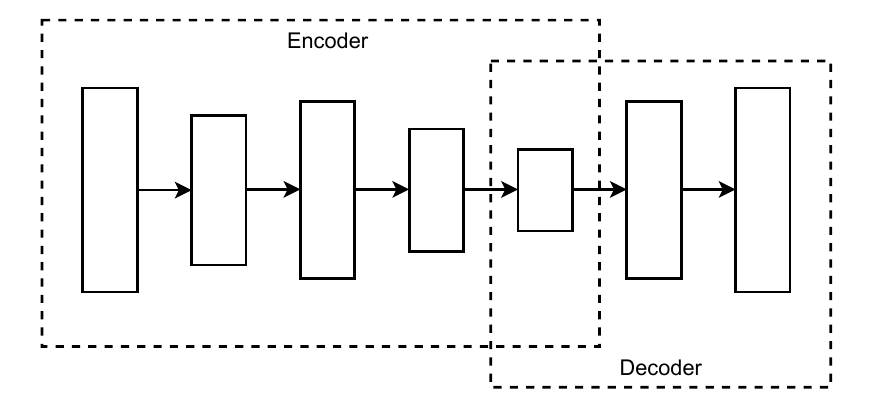}
    \caption{Example of an asymmetric design allowed by Assunção et al.~\cite{assuncao2018automatic}. The layer with the lowest dimensionality is the one that defines the representations.}
    \label{fig:assuncao_ae}
\end{figure}

Assunção et al.~\cite{assuncao2018automatic} also use a \gls{ga} with a similar representation, again restricted to dense layers. However, they adopt an asymmetric design, where both the encoder and decoder are evolved. In this case, the entire network is evolved without explicitly separating which layers belong to the encoder or decoder. The output layer of the encoder is selected as the one with the lowest dimensionality. An example of this asymmetric design is shown in Figure\ref{fig:assuncao_ae}.

Instead of using reconstruction loss as the fitness function, their approach minimises a composite objective based on three criteria: (i) the quality of the extracted features (measured by classification error in a downstream image classification task), (ii) the dimensionality of the extracted features, and (iii) the number of layers in the decoder.

Suganuma et al.~\cite{suganuma2018exploiting} use a (1+$\lambda$)-ES to evolve convolutional \glspl{ae}. To reduce the search space, the approach follows a symmetric design, where the decoder is a mirrored version of the encoder. The evolutionary process is restricted to convolutional layers, and both layer dimensionality and kernel size are encoded, with values selected from a predefined set. However, the evolutionary process also encodes the connectivity between layers, offering greater flexibility beyond traditional sequential architectures.

Similarly, Hajewski et al.~\cite{hajewski2020evolving, hajewski2021distributed} evolve convolutional \glspl{ae} using a ($\mu$+$\lambda$)-ES. The search space includes the number of layers and their hyperparameters, such as layer dimensionality and kernel size. Individuals are trained in a distributed environment, and fitness is computed using reconstruction loss on unseen data. Their method adopts a cell-based search space, in which only the layers inside each cell are evolved. The best-performing cell is later integrated into a fixed outer skeleton architecture.

Han et al.~\cite{han2022autonas} propose a \gls{ga} as part of a \gls{nas} framework that evolves kernel sizes in convolutional layers. Each kernel configuration is mapped to an integer, and each convolutional layer has its own assigned value. Individuals are sampled from a supernet that encodes the full search space, and only the weights of affected layers are updated during training. Evaluation is based on reconstruction loss, and variation is introduced via crossover and mutation.

Sun et al.~\cite{sun2018particle} use a \gls{pso} to evolve convolutional \glspl{ae}, guided by reconstruction loss. The approach evolves the encoder using a variable-length evolutionary representation, while the decoder is mirrored. The genotype encodes convolutional and pooling layers, along with their associated hyperparameters. Features extracted from the best individual are used in a downstream image classification task. The authors also study the impact of varying amounts of labelled data on downstream performance.

In a similar line, Kanwal et al.~\cite{kanwal2021evolving} propose a \gls{pso} to search for the optimal convolutional \gls{ae} topology, using a multi-objective fitness function. The objectives include (i) reconstruction loss on the pretext task, (ii) classification accuracy on a downstream image task, and (iii) the number of training parameters.

Finally, Dimanov et al.~\cite{dimanov2021moncae} introduce MONCAE, a multi-objective \gls{ec} algorithm that evolves convolutional \glspl{ae} based on reconstruction error and compression ability. After evolution, an additional fine-tuning phase is applied by training the final population for more epochs. Individuals exceeding a predefined compression threshold are then evaluated in a downstream image classification task using three datasets: MNIST, F-MNIST, and CIFAR-10.

Several studies in the literature address the evolution of neural network topologies for variational \glspl{ae}. Hajewski and Oliveira~\cite{hajewski2020evolutionary, hajewski2021efficient} evolve the number and size of dense layers using a ($\mu$+$\lambda$)-ES. Their approach allows for asymmetric \glspl{ae}, where the decoder is evolved independently of the encoder, without mirroring. This design increases the size of the search space, which in turn requires more evaluations for convergence. To reduce computational cost, candidate solutions are trained on a subset of the dataset with an early stopping mechanism.

\begin{figure}[!ht]
    \centering
    \includegraphics[width=1.\linewidth]{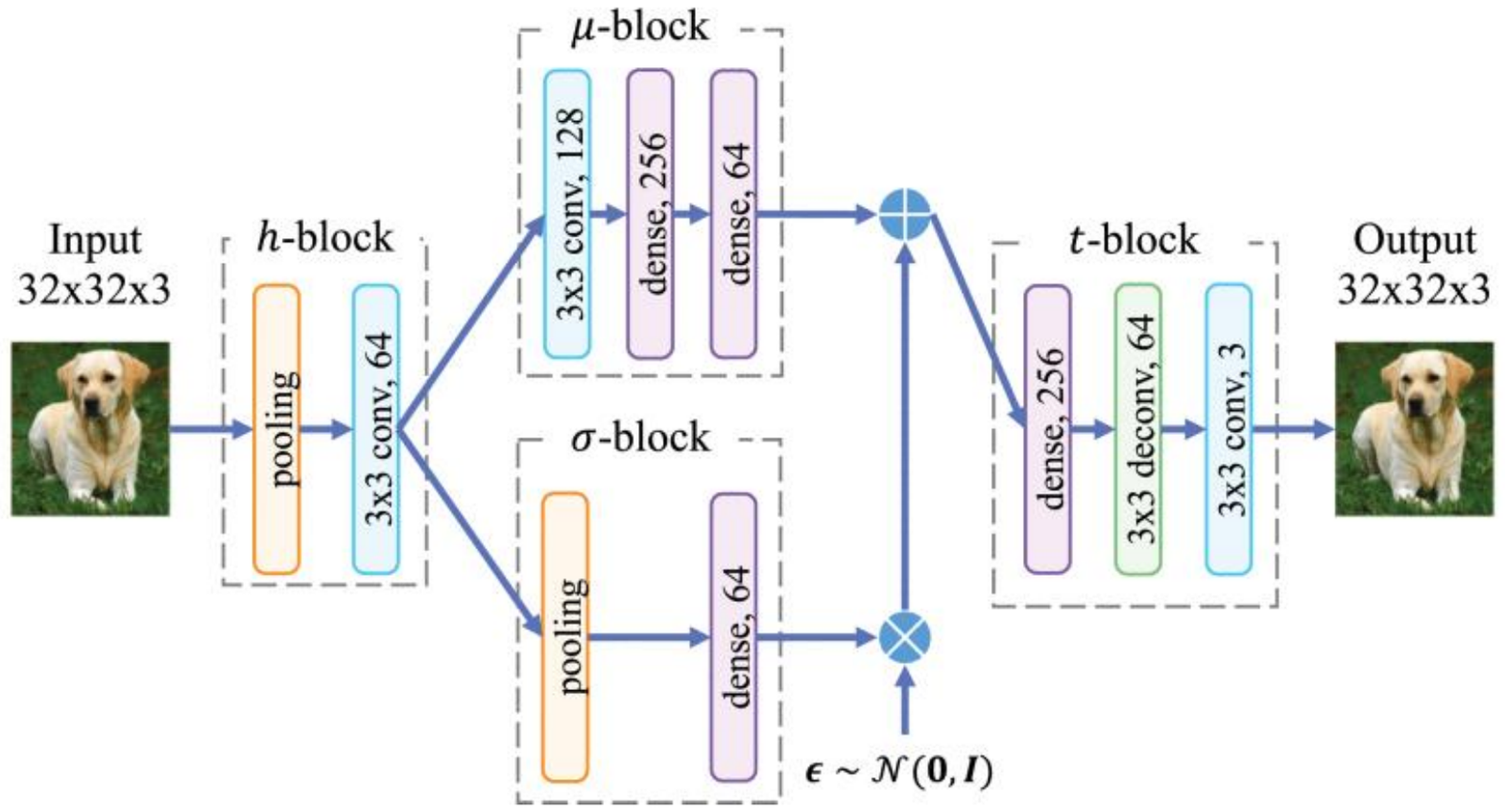}
    \caption{Skeleton of asymmetrical convolutional VAE used by Chen et al.~\cite{chen2020evolving}.}
    \label{fig:evovae}
\end{figure}

Chen et al.~\cite{chen2020evolving} propose EvoVAE, a method that uses a variable-length \gls{ga} to evolve convolutional variational \glspl{ae}. The algorithm encodes four types of layers -- dense, convolution, pooling, and deconvolution -- along with their respective hyperparameters. A distinctive feature of EvoVAE is its genotype structure, which is divided into four blocks, as shown in Figure\ref{fig:evovae}. The h-block serves as a shared backbone on the encoder side. The $\mu$ and $\sigma$ blocks are separate, each connecting to the backbone, and are used to model the latent space. The t-block functions as the decoder.

An implication from EvoVAE's design choices is that the crossover operator only allows exchange of information within the same block. Additionally, due to the variable-length genotype, a prior alignment of chromosomes is required before crossover can be applied.

Shang et al.~\cite{shang2024efficient} identified two main limitations in EvoVAE: a large search space and limited exploration capability of the crossover operator. To address the first issue, their proposed algorithm, AOC-VAE, alternates between evolving the encoder and the decoder every m generations. To improve exploration, they introduce an adaptive crossover operator in which alignment between selected individuals depends on the longest shared subsequences identified in the parent genotypes.

Both EvoVAE and AOC-VAE evaluate the learned representations in downstream image classification tasks and assess how performance is affected by the number of labelled samples available.

All previously mentioned works rely on \glspl{ae} trained for a reconstruction task that, along with any additional objectives, typically minimise the L2 loss. However, other types of architectures and pretext objectives can also be used for representation learning.

For example, Zhang et al.~\cite{zhang2021neural} evolve deep neural network topologies by performing representation learning on randomly assigned labels. These labels follow a discrete uniform distribution, with the number of classes matching that of the ground-truth dataset. The pretext task involves training an overparameterised supernet on the modified dataset. The evolutionary process then searches for the best topologies within the supernet’s subspace. To evaluate fitness, the authors introduce an angle-based metric that measures the distance between initial and trained weights, aiming to favour subnets that converge more quickly. The best individuals are retrained on the original dataset for the downstream task.

Xue et al.~\cite{xue2021fast} evolve new architectures from an existing overparameterised network that supports multiple layer options. Mutations are retained only if they result in lower validation loss, which is computed through contrastive \gls{ssl}.

In a similar direction, MaskTAS~\cite{yan2024masked} evolves vision transformer architectures within a supernet-constrained search space. Prior to the \gls{nas} stage, a supernet is trained using masked image modelling and self-distillation. In the first step, a teacher supernet is trained on masked image patches, following the approach of He et al.~\cite{he2022masked} (see Section\ref{sec:ssl}). Then, a student supernet is trained by sampling subnets from the supernet at each batch. These subnets are updated using both the masking task and a prediction loss based on latent representations from the teacher.

During the \gls{nas} phase, an \gls{ec} algorithm samples subnets from the student supernet and evolves them. Individuals are evaluated using a similarity metric that compares their learned representations to those from the teacher. The top k individuals are selected as parents, and variation is introduced through crossover and mutation. This process continues until a sufficient number of valid individuals -- meeting supernet constraints -- are generated. The best individual is then fine-tuned to obtain final performance results.

Garcia et al.~\cite{garcia2024progressive} evolve convolutional neural network topologies based on performance in the rotation prediction task proposed by Gidaris et al.~\cite{komodakis2018unsupervised}. Their method progressively evolves convolutional blocks with varying hyperparameters and connections using Cartesian \gls{gp}. The search space is constrained within a predefined outer skeleton composed of normal and reduction blocks. Normal blocks are evolved, while reduction blocks are fixed and include pooling layers. The best individual is retrained from scratch on the full downstream task, an image classification problem.

\begin{figure}[!ht]
    \centering
    \includegraphics[width=.8\linewidth]{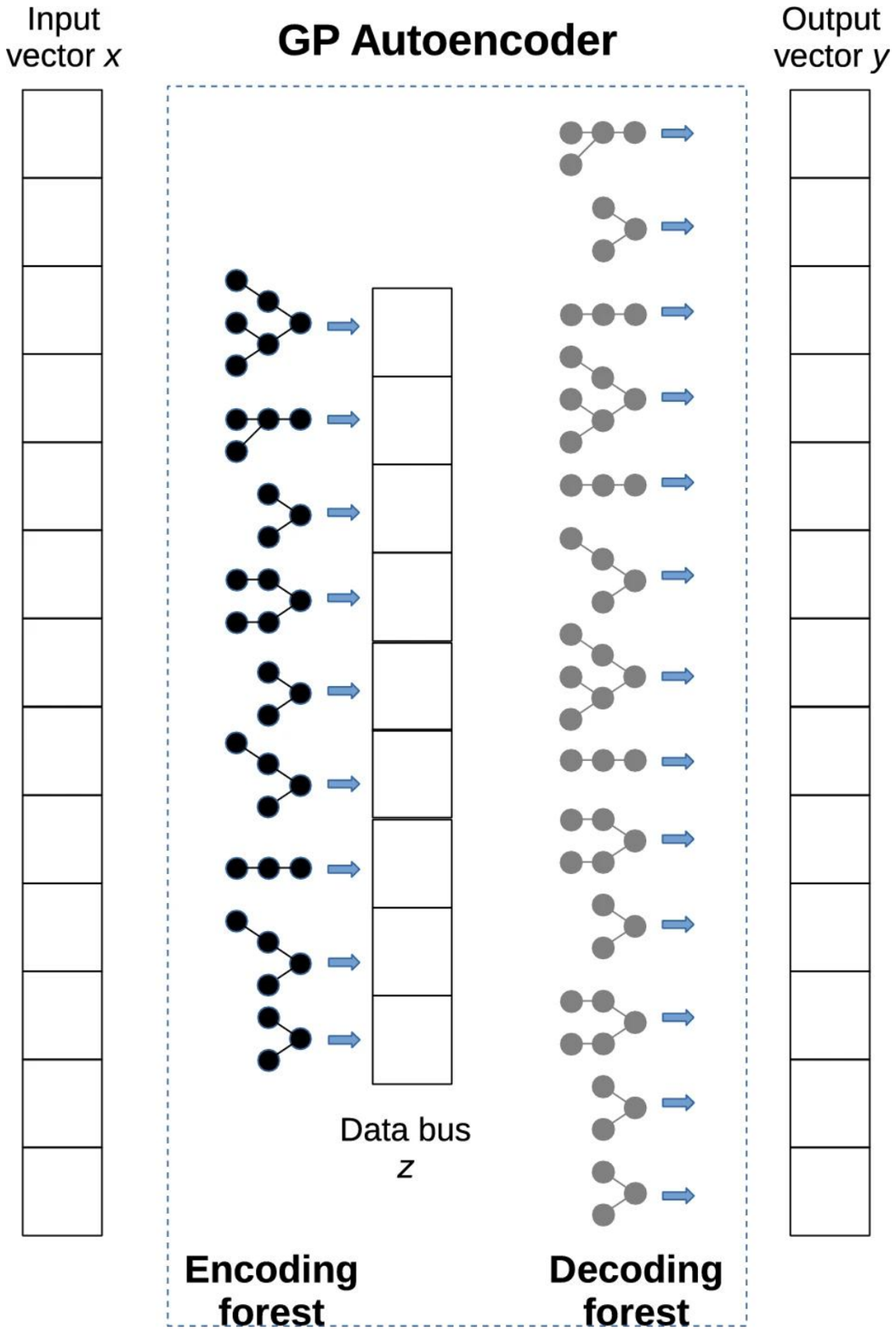}
    \caption{Representation of GP based autoencoder~\cite{rodriguez2019evolving}. Each tree uses a subset of its input to produce a single value of the representation (encoding forest), or the output vector $y$ (decoding forest).}
    \label{fig:gp_ae}
\end{figure}

Topologies do not necessarily need to be neural networks. Structures evolved from some evolutionary process can behave as \gls{ml} models and replace deep neural networks. Within this spectrum, Rodriguez-Coayahuitl et al.~\cite{rodriguez2019evolving} evolve two forest of \gls{gp} structures. Two populations are maintained, one to encode the inputs and the other to decode the output. At the macro level, the algorithm behaves like a \gls{ga} and each element from the \gls{ga} contains a tree-based representation. The overview of their \gls{ae} is depicted in figure~\ref{fig:gp_ae}.

Given an input with $n$ dimensions and a representation of dimensionality $m$, the encoder forest consists of $m$ trees, while the decoder contains $n$ trees. To handle high-dimensional problems, the leaves of each tree are restricted to a contiguous subset of features. In the encoder, each tree operates on a portion of the input vector $x$. In the decoder, the representation is divided into subsets, referred to as the data bus $z$ in Figure~\ref{fig:gp_ae}. Variation operators are applied at a macro level, using single-tree or single-point crossover, along with a mutation operator that randomly generates new trees. Schofield et al.~\cite{schofield2023genetic} note that, due to this subset-based design, input reconstruction is performed by considering only the features within the same subset. Their approach replaces only the encoder component of the \gls{ae} and allows any part of the input to be mapped to any feature in the learned representation.
Individuals are evaluated using \gls{mse} and evolved through the All Index Crossover operator and standard \gls{gp} mutation.

\subsubsection{Pretext task -- Learning}

The earliest approaches in this category design \gls{ec} algorithms to evolve the weights of an \gls{ae}, positioning them as an alternative to gradient descent and removing the requirement for differentiable neural network architectures.

David and Greental~\cite{david2014genetic} evolve the weights of an \gls{ae} with a fixed architecture. The encoder and decoder weights are tied, which substantially reduces the search space. Since the \gls{ae} is applied to an image-based task, the fitness function operates in the pixel space, using the inverse of the \gls{mse} as the evaluation metric. While the best individuals are further trained using backpropagation, the worst ones are replaced by the top-performing offspring. Offspring are generated using crossover (by exchanging weights between parents) and mutation (by randomly setting weights to zero).

To evolve weights in more complex \gls{ae} structures without excessively increasing the search space, Cai et al.~\cite{cai2018novel} propose a block-based search strategy, where each block evolves the weights of stacked \glspl{ae}. In the first block, the weights are evolved to reconstruct the raw inputs. In the subsequent blocks, each new \gls{ae} is evolved to reconstruct the inputs based on previously learned representations, meaning the decoder is only used when evolving its corresponding block.

At each block, the evolutionary process is guided by the Non-dominated Sorting Genetic Algorithm II (NSGA-II). Two objectives are considered: minimising reconstruction error and maximising the L1-norm (the sum of absolute weights). These conflicting goals promote sparsity in the learned weights while helping to prevent overfitting.

Evolving weights directly provides fine-grained control over what is being optimised but comes at the cost of a significantly larger search space. An alternative is to evolve the hyperparameters used in the loss function instead of the weights themselves.

For example, Cheng et al.~\cite{cheng2020adaptive} evolve hyperparameters for sparse \glspl{ae}. Their approach does not enforce the same sparsity penalty coefficient for all neurons. Instead, each neuron can have its own coefficient. The authors propose an \gls{ec} algorithm that evolves both the set of individual penalty coefficients and the learning rate $\lambda$. The evolutionary process is guided by two objectives derived from the loss function: one minimises the reconstruction loss, and the other minimises the sparsity penalty term. The method is evaluated on multiple network architectures using an image classification downstream task. Additionally, an ablation study is conducted to assess the impact of the number of labelled samples used in downstream training.

Wu et al.~\cite{wu2024evae} and Emm and Zhang~\cite{emm2024self} propose eVAE and SA-eInfoVAE, respectively. In general, the loss function for a variational \gls{ae} consists of two components: one for minimising reconstruction loss, and another for minimising the Kullback–Leibler (KL) divergence between the prior distribution and the distribution learned by the encoder.

In eVAE, a coefficient $\beta$ is introduced to control the relative importance of the KL divergence term, and this value is evolved. The evolutionary process operates by first training a variational \gls{ae} for $N$ epochs. Then, a population of candidate $\beta$ values is generated. These are evaluated based on the direction of the stochastic gradient and are changed using `variational' crossover and mutation. The best $\beta$ value is selected and returned to eVAE, initiating another round of training.

SA-eInfoVAE considers a modified loss function that includes an additional KL divergence term between the prior distribution and the distribution of sampled latent vectors. This term is associated with a second coefficient, $\lambda$. While eVAE evolves $\beta$, SA-eInfoVAE evolves both $\beta$ and $\lambda$ independently. In addition, the crossover operator used in SA-eInfoVAE is self-adaptive, dynamically selecting which portions of the parents are inherited.

Sors et al.~\cite{sors2021simple} explore the decomposition of contrastive loss into multiple sub-losses, each associated with its own coefficient. These coefficients are evolved to maximise the mean average precision (mAP) of the learned representations. Although their main optimisation method is based on gradient descent, the authors also use CMA-ES as a baseline evolutionary approach.

A more recent trend in representation learning focuses on reducing model footprint and improving inference-time efficiency. One common strategy is quantisation -- a technique that reduces the precision of weights while attempting to maintain model performance. Quantisation is parameterised by two factors: the bit width used to represent a value, and a scale factor that maps the original range of values to a more compact domain of quantised values.

EvolQ~\cite{frumkin2023jumping} is a post-training quantisation algorithm that compresses the weights of a vision transformer, guided by a contrastive \gls{ssl} loss. The first step of EvolQ involves learning the scale factors applied to weight vectors and activations. The evolutionary process is applied separately to each transformer block. Within each block, the evolutionary process generates a population of perturbation vectors that modify the learned scales. These vectors are evolved to find the quantised configuration that minimises the contrastive loss.

CLAMP-ViT~\cite{ramachandran2024clamp} extends this work by evolving not only the scale parameters but also the bit width itself. Ramachandran et al.~\cite{ramachandran2024algorithm} propose a similar mechanism tailored for weights represented using Logarithmic Posits (LP), which have shown benefits over standard floating-point formats for neural network inference~\cite{gustafson2017beating}. LP types are parameterised by four coefficients, which are evolved using a fitness function that balances two objectives: layer-wise contrastive loss and bit width.

Another application of \gls{ec} in \gls{ssl} is the discovery of pretext tasks that lead to better representations. Previous work has shown that the effectiveness of a pretext task can vary depending on the downstream task~\cite{li2022does}. As such, combining multiple pretext tasks may result in more robust and generalisable representations.

Jin et al.~\cite{jin2021automated} propose AutoSSL, a framework designed for graph neural networks. It combines five graph-based pretext tasks into a single loss function using a linear combination. The coefficients associated with each pretext task are evolved using CMA-ES.

\begin{figure}[!ht]
    \centering
    \includegraphics[width=.9\linewidth]{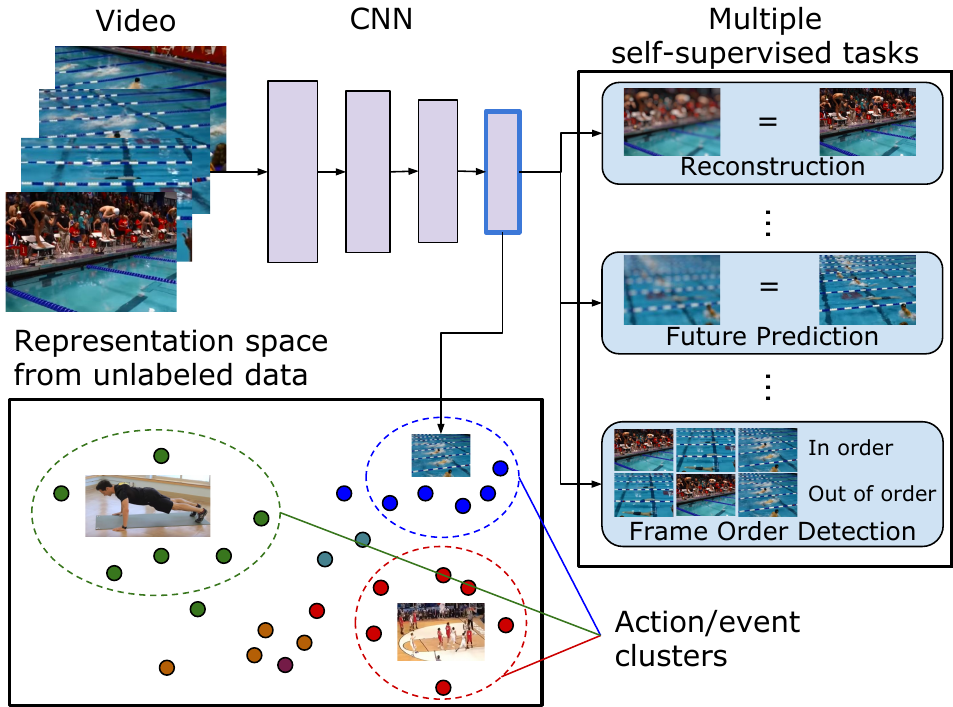}
    \caption{Overview of ELo~\cite{piergiovanni2020evolving}.}
    \label{fig:elo}
\end{figure}

Similarly, Piergiovanni et al. introduce Evolving Losses (ELo), which targets video representation learning by combining tasks from different modalities. The final loss function integrates both single-modality and cross-modality distillation losses. Coefficients for each task are evolved using CMA-ES, guided by an unsupervised metric that applies K-Means to the learned representations and evaluates how well the resulting clusters follow Zipf’s law. An overview of the ELo framework is shown in Figure~\ref{fig:elo}.

One notable aspect of the ELo experiments is the detailed analysis of how the number of labelled samples affects downstream task performance. Remarkably, ELo achieves results comparable to a state-of-the-art supervised model on the Kinetics-400 activity recognition dataset, using only half the labelled data.

Liu et al.~\cite{liu2024collaborative} apply an \gls{es} to learn representations for an image classification task. Individuals are encoded as vectors of weights, each associated with a specific pretext task used for representation learning. To encourage population diversity, some genes are initialised with a weight of zero. This design choice also simplifies evaluation, as pretext tasks with a weight of zero are not executed.

The fitness of each individual reflects the performance of the corresponding weight combination on the downstream task. After the evolutionary process concludes, the weights of the best individual are further refined through extended training on the downstream task before evaluating the learned representations on unseen datasets.

\subsubsection{Pretext task -- Multiple components}

Several of the surveyed studies use \gls{ec} to evolve multiple components simultaneously, among the three previously identified in this section. These works recognise that, although each component can individually contribute to the overall performance in solving a \gls{ml} task, interdependencies between components can hinder the search for an optimal combination. While earlier approaches typically optimise a single component while keeping the others fixed, these studies adopt a more holistic perspective. However, this broader view comes with the drawback of significantly expanding the search space, thereby increasing the difficulty of the optimisation task.

Most of the studies identified in this group focus on evolving both topological and learning-related aspects. Lander and Shang propose EvoAE~\cite{lander2015evoae}, a framework that offers greater flexibility in the evolutionary process by targeting both the weight space and the network structure. To manage this flexibility, the same architecture is imposed on both the encoder and decoder. For scalability, individuals are trained on a data subset, and a post-training phase is applied to the best-performing \gls{ae}, which is then evaluated on previously unseen data.

Charte et al.~\cite{charte2019automating, charte2020evoaaa} explore the use of several \gls{ec} algorithms to evolve \glspl{ae}. Their approach uses a fixed-length genotype, where genes encode topological features such as the type of \gls{ae}, the number of layers, and the size and activation function of each layer, assuming a symmetric topology. The final gene specifies the loss function used during training. The fitness function combines training loss, the number of layers, and the dimensionality of the learned representations. The authors evaluate this setup using \gls{ga}, \gls{es}, and \gls{de} algorithms.

Some studies also evolve dataset-related components alongside other elements. Li et al.~\cite{li2022does} propose an evolutionary approach that combines ideas from ELo~\cite{piergiovanni2020evolving} and the data augmentation parameter search from Barrett et al.~\cite{barrett2023evolutionary} to explore the impact of \gls{ssl} in reinforcement learning from pixels. A \gls{pso} is used to evolve a set of coefficients $w_{i},\ i \in {1, \dots, N}$, where $N$ is the number of self-supervised tasks, along with magnitude parameters $m_{j},\ j \in {1,2}$, which control the strength of data augmentation. Two magnitude parameters are used to support \gls{ssl} algorithms based on dual-branch architectures.

Li et al.~\cite{li2023extensible} extend GenNAS by expanding the flexibility in the search for synthetic labels. As shown earlier in Figure\ref{fig:gennas}, a convolutional layer is attached to the output of each network stage. Their extension enlarges the search space by evolving not only the parameters that generate the synthetic labels (dataset), but also the convolutional hyperparameters (topology), along with the learning rate and weight initialisation method (learning).

Preen et al.~\cite{preen2021autoencoding} explore the use of a Learning Classifier System to adaptively decompose the input domain into a population of smaller \glspl{ae}. Each \gls{ae} is evolved by modifying both structural and learning components through mutation. These include changes to connectivity, the number of neurons (topology), weights, and learning rate (learning). Fitness is based on downstream performance, defined as the average probability of a classifier matching a rule specified a priori.

Silhan et al.~\cite{silhan2019evolution} propose a layer-wise evolutionary strategy for \glspl{ae}, where a new layer is added to both the encoder and decoder at each step to maintain symmetry. The evolutionary process evolves two topological parameters -- dropout ratio and activation function -- and two learning parameters -- learning rate and momentum. At each generation, these hyperparameters are mutated, and weights are inherited by offspring, following a Lamarckian evolution strategy. Each individual is evaluated using the Local Continuity Meta-Criterion (LCMC)~\cite{chen2009local}, a metric that measures the quality of dimensionality reduction.

Outside the scope of \glspl{ae}, Sun et al.~\cite{sun2018evolving} propose EuDNN, a method to evolve both connection weights and activation functions. A layer-wise mechanism is used, in which evolution occurs within the input subspace at each layer. To encode individual weights, a set of basis vectors is linearly combined with a bias vector to generate an output vector $a$. Then, $n$ output vectors orthogonal to $a$ are generated. The genotype encodes a subset of these orthogonal vectors, the corresponding bias values, and an index representing the activation function.

Once representations are learned, a support vector machine (SVM) is trained on them, allowing fine-tuning of the evolved weights. The accuracy of the SVM is used as the fitness signal to guide the evolutionary process.

Vinhas et al.~\cite{vinhas2024towards} present EvoDeNSS, a framework that evolves both topological and learning aspects of neural networks. The search space is defined a priori using a context-free grammar, which biases the evolutionary process toward more promising solutions. The topological components include the number of layers, layer types, and their hyperparameters. On the learning side, the evolved elements include the learning rate, number of training epochs, batch size, optimiser, and its hyperparameters.

\begin{figure}[!ht]
    \centering
    \begin{tikzpicture}[node distance=0pt]
        
        \small
        \node[align=left] (placeholder) {};
        \node[align=center, right=of placeholder, xshift=1.25cm] (outer) {outer-level};
        \scriptsize\smaller[1]
        \node[draw, node distance=0.2cm, text width=2cm, minimum height=1cm, align=center, below=of placeholder] (features1) {$<$features$>$};
        \node[draw, text width=2cm, minimum height=1cm, align=center, right=of features1] (features2) {$<$features$>$};
        \node[draw, text width=2cm, text depth=-0.0cm, minimum height=1cm, align=center, right=of features2] (classification) {$<$classification$>$};
        
        \small
        \node[align=center, node distance=3.2cm, below=of placeholder, xshift=2.98cm] (inner) {inner-level}; 
        \scriptsize\smaller[1]
        \node[align=center, node distance=2.5cm, below=of features2, xshift=-2.5cm] (features_text) {$<$features$>$};
        \node[draw, node distance=0.1cm, text width=1.5cm, text depth=1.5cm, minimum height=2cm, align=left, below=of features_text] (_features1) {\{DSGE:1\},\\\{\}};
        \node[draw, text width=1.9cm, text depth=1.5cm, minimum height=2cm, align=left, right=of _features1] (conv) {\{DSGE:0\},\\\{out\_channels:93\\~kernel\_size:3\\~stride:3\}};
        \node[align=center, node distance=0.1cm, above=of conv] (conv_text) {$<$convolution$>$};
        \node[draw, text width=1.5cm, text depth=1.5cm, minimum height=2cm, align=left, right=of conv] (padding) {\{DSGE:1\},\\\{\}};
        \node[align=center, node distance=0.06cm, above=of padding] (padding_text) {$<$padding$>$};
        \node[draw, text width=1.5cm, text depth=1.5cm, minimum height=2cm, align=left, right=of padding] (act_function) {\{DSGE:1\},\\\{\}};
        \node[align=center, node distance=0.1cm, above=of act_function] (act_function_text) {$<$act\_function$>$};
        \node[draw, text width=1.5cm, text depth=1.5cm, minimum height=2cm, align=left, right=of act_function] (bias) {\{DSGE:1\},\\\{\}};
        \node[align=center, node distance=0.1cm, above=of bias] (bias_text) {$<$bias$>$};
        
        \draw [densely dotted] (features2.south west) -- ([yshift=0.3cm]_features1.north west);
        \draw [densely dotted] (features2.south east) -- ([yshift=0.3cm]bias.north east);

    \end{tikzpicture}
    \caption{Bi-level representation used in EvoDeNSS~\cite{vinhas2024towards}.}
    \label{fig:evodenss}
\end{figure}
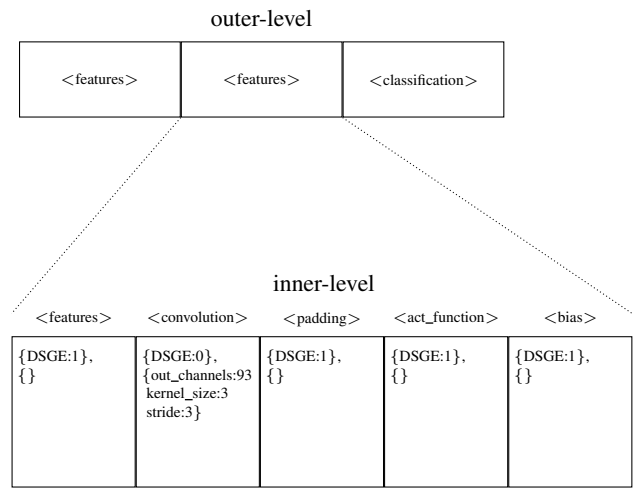

EvoDeNSS employs a bi-level evolutionary representation. At the outer level, it follows a \gls{ga} structure consisting of an array of macro blocks, where each block can represent either a network layer or a learning-related component. At the inner level, each block encodes its specific hyperparameters using Dynamic Structured Grammar Evolution (DSGE). An example of the genotype structure is shown in Figure~\ref{fig:evodenss}.

Evolved networks learn representations using Barlow Twins. These representations are then used to train a dense layer for an image classification downstream task, with fitness being measured by accuracy in the downstream task. The authors also evaluate the impact of using a dataset with a scarce number of labelled data points in the evolutionary process, comparing both \gls{ssl} and Supervised Learning scenarios.

\subsubsection{Downstream task}
Some studies address the use of \gls{ec} on top of pretrained models in the context of a given self-supervised pretext task. Similar to the pretext task, \gls{ec} can target the evolution of dataset, learning, or topology components that are related to the downstream task.

Steinmetz et al.~\cite{steinmetz2024st} tackle a style transfer task in the context of audio production. They propose that a series of audio effects can be applied to an input audio signal to approximate a reference audio, where each effect is represented as a parametrised function. The authors first train a self-supervised model to learn audio representations. The downstream task optimises the parameters of the audio effects using CMA-ES. Fitness is evaluated by calculating the cosine similarity between the embeddings of the reference audio and the input audio after applying the audio effects chain.

Zhou et al.~\cite{zhou2024legal} fine-tune a self-supervised pretrained model for a text matching task using supervised contrastive learning. The core idea is to ensure that pairs of matching texts are represented similarly in the latent space, while non-matching pairs are pushed apart. Based on the misclassified examples from the trained model, an \gls{ec} algorithm is applied to evolve alternative options that match the original text. These new pairs are added to the original set, and another round of contrastive training is performed.

Sun et al.~\cite{sun2023multitask} also target the evolution of inputs in the downstream task. They use pretrained self-supervised models for prompt-based learning based on pretrained prompts. Their approach is based on the idea that converting classification problems into a masked language modelling problem makes it easy to adapt to multiple downstream tasks. They argue that the pretrained model trained during the pretext task does not need to be modified if prompt tuning is performed. They decompose the downstream fine-tuning process into two stages to bridge the gap between pretext and downstream tasks. In the first stage, a binary matrix $K*T$ is optimised via Bayesian optimisation to determine which prompts should be activated for each task. In the second stage, CMA-ES is used to evolve prompts that enhance the performance of the downstream tasks they are activated for.

In contrast to the previous work by Sun et al., Bu et al.~\cite{bu2024layer} adopt an opposing view, arguing that there is no fixed, human-designed set of learning parameters that consistently outperforms others across a range of NLP tasks. They propose AutoFT, a method to fine-tune pretrained BERT representations for specific downstream tasks by evolving learning parameters. AutoFT performs combinatorial optimisation of different parameter sets via evolution. A \gls{ga} maintains a population of individuals, each representing a parameter set for each layer of the final network (including the pretrained BERT layers). This approach has shown slightly better results than using a fixed parameter set for fine-tuning.

Shen et al.~\cite{shen2024evolutionary} employ a network that learns image representations based on transformations such as rotations or different colour arrangements from visible light and infrared images, which are then fused. On top of the learned representations, a \gls{pso} algorithm is applied to evolve learning hyperparameters used during gradient descent, including learning rate, number of epochs, and batch size.

Hu et al.~\cite{hu2024joint} use a set of stacked models on top of a pretrained BERT model for Named Entity Recognition (NER). Specifically, the last layer is a \gls{brb}~\cite{yang2006belief}, and its parameters are optimised using a distributed CMA-ES. The role of the \gls{brb} is to filter noise and ensure accurate entity classification by establishing regular expressions as rules.

In terms of using \gls{ec} to target network topology at the downstream stage, Ludwig and Claes~\cite{ludwig2023compressing} focus on pruning self-supervised wav2vec models while maintaining generalisation ability, by adjusting the connectivity between layers. Their motivation is that original wav2vec models have a considerable footprint, which impacts latency. Wav2vec models consist of 24 blocks, each with 2 fully connected layers that account for a large portion of the encoder parameters. Since the weights of fully connected layers can be represented as matrices, the authors use a \gls{ga} that encodes groups of consecutive matrix rows for each fully connected layer and evolves which groups should be pruned, guided by the word error rate. The pruned models are then fine-tuned with the same data used during the evolution stage.

Instead of encoding parameterised topology aspects in the evolutionary process, \gls{ec} can be used directly to solve the downstream task. For example, Rodrigues et al.~\cite{rodrigues2023performance} apply evolved individuals on top of a set of representations trained in a self-supervised manner. They use \gls{gp} to evolve models that solve downstream tasks in the tabular data domain. The authors note that while \gls{ssl} works seamlessly for structured domains such as NLP and computer vision, it is less effective for tabular data due to its heterogeneity and lack of global structure. Indeed, they show that incorporating \gls{ssl} representations actually hinders the performance of downstream models.

Finally, Ha and Gao~\cite{ha2021evolving} apply evolution not only to the topological aspects of the downstream task but also to those of the pretext task. However, it should be noted that evolving topological components for both the pretext and downstream tasks has implications for the search space and convergence speed. Their system architecture consists of three different types of textual embeddings, each submitted to one of three \glspl{ae}. The representations learned by these \glspl{ae} are concatenated and passed to a classifier layer for text classification. The evolutionary system is a \gls{ga}, with individual index values mapping to specific layer types, layer hyperparameters, or layer deactivation. Both the \glspl{ae} and the classifier layer are evolved based on performance on a validation set.

\subsection{SSL for EC}
\label{subsec:ssl_for_ec}

As previously discussed in Section~\ref{sec:eml}, \gls{ssl} models can be integrated into \gls{ec} by targeting any of the components of an evolutionary algorithm to enhance aspects of the evolution process. This area of research is rapidly expanding, with several works focusing on three key components of an evolutionary algorithm: evolutionary representation, variation operators and fitness evaluation. The studies that fall under this category are described in Table~\ref{tab:works_ssl_for_ec}. It describes what does a certain work evolve, if it uses an off-the-shelf model in the evolution process or if tries to search new models, the type of goal defined in the pretext task, what part of the model is integrated in the evolutionary process, and what \gls{ec} component the model targets. 


\renewcommand{\arraystretch}{1}
\begin{sidewaystable}
\scriptsize
\centering
\begin{tabular}{>{\centering\arraybackslash}m{2.2cm}>{\centering\arraybackslash}m{3cm}>{\centering\arraybackslash}m{2cm}>{\centering\arraybackslash}m{2.5cm}>{\centering\arraybackslash}m{2cm}>{\centering\arraybackslash}m{3cm}}\toprule
 & & \multicolumn{4}{c}{\textbf{SSL engagement with EC}} \\\cmidrule{3-6}
\textbf{Work name} & \textbf{What is evolved} & \textbf{Off-the-shelf model} & \textbf{SSL goal type} & \textbf{What is used for evolution} & \textbf{EC \break component targeted} \\
\hline
Wei et al. \cite{wei2021self} & Neural networks & No & Regression + Contrastive (separately) & Downstream output & Fitness evaluation \\
\hline
CLIP MasterPrints \cite{freiberger2024fooling} & Image representations & Yes & Multimodal agreement & Pretext output & Fitness evaluation \\
\hline
EvoProteus \cite{sacadura2024design} & Image generation parameters & Yes & Multimodal agreement & Pretext output & Fitness evaluation \\
\hline
Machado et al. \cite{machado2024pixels} & Prompts & Yes & Multimodal agreement & Pretext output & Fitness evaluation \\
\hline
BPT-VLM \cite{yu2023black} & Prompts & Yes & Multimodal agreement & Pretext output & Fitness evaluation \\
\hline
HEP-ENAS \cite{xiao2024evolutionary} & Neural networks & No & Input \break reconstruction & Representations & Genotype-Phenotype mapping \\
\hline
Hu et al. \cite{hu2024autoencoder} & Several optimisation problems & No & Input \break reconstruction & Representations & Genotype-Phenotype mapping \\
\hline
DAE-GP \cite{wittenberg2020dae, wittenberg2023denoising, wittenberg2022denoising, reiter2023pretraining} & Several optimisation problems & No & Input \break reconstruction & Pretext output & Operators \\
\hline
EAEPSO \cite{yuan2023particle} & Neural networks & No & Input \break reconstruction & Representations & Genotype-Phenotype mapping \\
\hline
MOEA/PSL \cite{tian2020solving} & Several optimisation problems & No & Input \break reconstruction & Representations & Genotype-Phenotype mapping \\
\hline
TTLBO \cite{cui2022bi} & Several optimisation problems & No & Input \break reconstruction & Representations & Genotype-Phenotype mapping \\
\hline
ENAO \cite{li2024enao} & Neural networks & No & Input \break reconstruction & Representations & Genotype Phenotype mapping \\
\hline
Gong et al. \cite{gong2024bridge} & Neural networks & No & Input \break reconstruction & Representations & Genotype-Phenotype mapping \\
\hline
Shem-Tov et al. \cite{shem2024deep} & Regression coefficients & No & Masking & Pretext task & Operators \\
\hline
DAGA \cite{churchill2014denoising} & Several optimisation problems & No & Input \break reconstruction & Pretext output & Operators \\
\hline
DAE-EDA \cite{probst2015denoising} & Several optimisation problems & No & Input \break reconstruction & Pretext output & Operators \\
\hline
Thakkar et al. \cite{thakkar2019autoencoder} & Several optimisation problems & No & Input \break reconstruction & Pretext output & Operators \\
\hline
Dang et al. \cite{dang2024novel} & Neural networks & No & Input \break reconstruction & Downstream output & Fitness evolution \\
\hline

\end{tabular}
\label{tab:works_ssl_for_ec}
\caption{Works where \gls{ssl} is applied for \gls{ec}}
\end{sidewaystable}
\renewcommand{\arraystretch}{1}

\subsubsection{Evolutionary Representation}

In \gls{ec}, representation, along with variation operators, shapes the search space, directly impacting the success of finding the optimal solution. \gls{ssl} features learned during the pretext task can be used to encode the input space, thereby delegating the optimisation task to the latent space.

One advantage of this approach is related to the dimensionality of the problem. Latent spaces can be viewed as compressed versions of the input space. By performing evolution in the latent space, the dimensionality of the problem is reduced, making it easier to navigate the search space. For example, Tian et al.~\cite{tian2020solving} use a denoising \gls{ae} integrated into an evolutionary process designed for large-scale multiobjective optimisation problems. Before each generation, an \gls{ae} is trained using non-dominated individuals, which are corrupted by randomly setting some chromosome values to zero. The \gls{ae} encodes solutions into the latent space, where genetic operators are applied. The solutions are then decoded back into the original search space for evaluation.

Cui et al.~\cite{cui2022bi} frame an \gls{ec} algorithm as a cooperative co-evolution problem with two populations to balance exploration and exploitation. One population performs evolution in the original search space, representing exploitation, while the other undergoes dimensionality reduction through an \gls{ae} and applies genetic operators in the reduced search space. The populations are split at the start of each generation, with the individuals with the worst fitness being assigned to the population fed to the \gls{ae}. This population is classified as performing exploration. In the initial generations, evolution does not rely on the \gls{ae} since it is gathering individuals for training.

Hu et al.~\cite{hu2024autoencoder} apply differential evolution to solve large-scale problems. They use a pretrained \gls{ae} to reduce the problem’s dimensionality and employ a clustering algorithm for interaction analysis in the latent space. For each cluster, the individual with the highest non-dominant rank is selected as a parent.

The exploration of \glspl{ae} to target the evolutionary representation component in evolutionary \gls{nas} problems is gaining momentum. In this context, the motivation extends beyond addressing the curse of dimensionality, aiming to also learn the mapping between the phenotypic and genotypic spaces. Neural network architecture phenotypes are inherently discrete and can vary in length. As a result, some \gls{nas} approaches adopt variable-length representations and/or explore discrete search spaces. This approach helps avoid the need to design complex mechanisms for mapping the genotype to the phenotype in order to obtain the final solution.

However, using variable-length genotypes has the drawback of requiring a redesign of variation operators. If not carefully managed, this can lead to convergence towards sub-optimal regions of the search space. By employing \glspl{ae} to learn this mapping, the process of learning the connections between the genotype and phenotype (and vice versa) can be automated. Additionally, neural network architecture genotypes can maintain a fixed length in continuous space, addressing these issues while preserving the integrity of neural network architectures.

Yuan et al.~\cite{yuan2023particle} evolve dense blocks using \gls{pso}. The dense blocks are first converted into block vectors. Before the evolutionary process begins, an \gls{ae} is trained on pairs of randomly generated block vectors. The \gls{ae} is trained not only to minimise the reconstruction error of each architecture but also to ensure consistency between two similarity metrics for the pair of block vectors, comparing them both in the original and latent space. Since block vectors have a variable length within specified bounds, the \gls{ae} expects an input dimensionality equal to the maximum allowed length. If a block vector is shorter than the upper bound, it is padded with zeros. During the evaluation phase of evolution, each solution is decoded, trained with stochastic gradient descent, and evaluated based on accuracy on a validation set.

Gong et al.~\cite{gong2024bridge} use a similar \gls{ae} approach but apply a \gls{ga} to enable the use of standard crossover and mutation operations over candidate solutions. The preprocessing of the architectural data provided to the \gls{ae} differs as well. Instead of padding, the authors treat each candidate as a sequence of chromosomes. Therefore, both the encoder and decoder of the \gls{ae} are \glspl{gru}, a type of \gls{rnn}.

Similarly, Xiao et al.~\cite{xiao2024evolutionary} view neural networks as sequences and integrate a transformer-based variational \gls{ae} within an evolutionary \gls{nas} system. The optimisation of the latent space is performed with CMA-ES. Li et al. propose ENAO~\cite{li2024enao}, an evolutionary \gls{nas} system that frames neural networks as graphs. The authors incorporate a graph-based variational \gls{ae} to encode architectures into the latent space, integrating it into an evolutionary system tested with both \gls{pso} and CMA-ES.

\subsubsection{Operators}

Replacing standard crossover and mutation operators with alternatives can be achieved using \gls{ssl}. By taking advantage of the problem structure and training representations that capture relationships between variables, one can incorporate knowledge to design informed operators.

Denoising \glspl{ae} are particularly useful for this purpose, as there are reports of their application in \glspl{ga} to sample individuals~\cite{probst2015denoising, churchill2014denoising}. The basic approach involves using individuals to train the \gls{ae} at each generation, and then creating offspring by sampling from the trained \gls{ae}. The training set can be formed using selected individuals through tournament selection~\cite{probst2015denoising} or truncation~\cite{churchill2014denoising}, which controls whether a completely new population is sampled or only part of it. DAE-GP~\cite{wittenberg2020dae} is the most prominent operator in this area of research. It consists of an \gls{ae} with \glspl{lstm} for both encoding and decoding, integrated into a \gls{gp} algorithm. Since individuals are represented as trees, they are converted into linear sequences in prefix notation before being fed to the network. New individuals are generated by inputting a corrupted version of an individual into the network, and iteratively using the network’s output as a new input for the \gls{ae}. This process is repeated for several steps. Given the importance of balancing exploration and exploitation in successful searches, the authors also found that adjusting the corruption strength and the number of sampling steps is crucial for achieving this balance~\cite{wittenberg2023denoising}. DAE-GP was later extended for real-world symbolic regression problems~\cite{wittenberg2022denoising} and by using a pretrained \gls{ae}~\cite{reiter2023pretraining}. In this extension, the \gls{ae} is trained on a random population before evolution begins. At each generation, the pretrained network is loaded and refined, demonstrating its potential for reducing computational cost.

Thakkar et al.~\cite{thakkar2019autoencoder} also pretrain \glspl{ae} before evolution. They provide information about game levels to an \gls{ae} and use it as a mutation operator within an evolutionary algorithm designed for game level generation. The authors compared the use of vanilla \glspl{ae} and variational \glspl{ae}, concluding that mutation based on variational \glspl{ae} resulted in game levels with more detail.

Another approach to combining \gls{ssl} with \gls{ec} operators is presented by Shem-Tov et al.~\cite{shem2024deep}. In their work, \gls{ssl} is not used to build representations. Instead, they design a mutation operator inspired by the masked language modelling problem, which is solved by BERT during the pretext task (predicting masked tokens). This operator is designed for \gls{gp} and enhances point mutation. The idea is to mask nodes of the individual to be mutated and use reinforcement learning to select new nodes that maximise fitness, using fitness improvement as a reward signal.

\subsubsection{Fitness}

\gls{ssl} models can contribute to fitness operators by assisting in the fitness assignment process. The output of these models can either be used to calculate fitness or directly serve as the fitness measure. \gls{ssl} models for fitness assignment can be classified into two main categories. They can either replace the fitness module itself or act as surrogate models to mitigate the computational burden commonly associated with evolutionary algorithms during the evaluation stage.

The most common scenario involves using off-the-shelf \gls{ssl} models. However, some of the models used in this context have been trained for both pretext and downstream tasks. Although we include such models in this category, we are primarily interested in models that generate self-supervised representations without prior downstream training.

CLIP, pretrained on multimodal data, provides representations that can be used to compute fitness. For instance, Machado et al.~\cite{machado2024pixels} incorporate CLIP into an evolutionary process which evolves images that depict a coin side. Fitness is calculated by using CLIP to compute the cosine similarity between each image and a textual description provided a priori. Following the same rationale, Sacadura et al.~\cite{sacadura2024design} integrate CLIP into their evolutionary framework to evolve a set of parameters for visual artifact generation, which are then evaluated using CLIP. Their goal is to generate visual artifacts that resemble a given concept. Similarly, Freiberger et al.~\cite{freiberger2024fooling} use CLIP for the same purpose. However, the authors note that CLIP models are susceptible to exploitation and developed an evolutionary system capable of evolving what they call ``master images'' -- images that produce high similarity to a wide range of prompts, while being unrelated to those prompts to the human eye. In this system, individuals are representations that are later decoded to generate images, framing the discovery of master images as a black-box optimisation problem. This exploitation problem is not specific to architectures that learn through \gls{ssl} though, as this a problem previously identified and tackled in the Supervised Learning paradigm~\cite{nguyen2015fooled,correia2016xfaces}.

Yu et al.~\cite{yu2023black} also tackle a black-box optimisation problem for cases where gradient information is unavailable. Unlike previous work, they jointly optimise text and image prompts for zero-shot classification. Each individual consists of a pair of prompts (one for text and one for image), treated as a continuous vector. The evaluation procedure involves solving the classification task with the additional information from the prompts fed to each of the CLIP encoders. Fitness is derived from CLIP, as the system uses cross-entropy loss to guide evolution.

Considering \gls{ssl} models as surrogate models, Wei et al.~\cite{wei2021self} address label scarcity in predictive performance modelling. They propose two neural predictors that can learn in a self-supervised manner. The first predictor trains neural network representations via regression. Using a dual-branch network (with no shared weights), the pretext task involves predicting the normalised graph edit distance between pairs of neural network architectures. The second predictor learns representations via contrastive learning. The key idea is to feed a batch containing neural network architecture information. For each architecture $a$, the positive set consists of $a$ itself and the architecture from the batch with the minimum graph edit distance to $a$. The remaining elements in the batch form the negative set. These neural predictors are integrated into an evolutionary \gls{nas} system. After training on the downstream task, the predictive performance of each neural predictor is used as fitness. Their system is evaluated in three search spaces: NasBench101, NasBench201, and DARTS, and compared against other \gls{nas} approaches based on gradients, Bayesian methods, and reinforcement learning. The authors demonstrate that their system achieves comparable performance to others using either neural predictor. Additionally, the search speed on DARTS is comparable to non-evolutionary approaches, countering the common argument that evolutionary \gls{nas} systems are time-consuming. However, the authors note that their contrastive learning predictor outperforms the regression-based one when search spaces are larger. Following a similar approach, although previously mentioned, it is worth noting that the output of GenNAS~\cite{li2021generic,li2023extensible} (and its optimal synthetic labels) is also integrated into evolutionary \gls{nas} algorithms to guide evolution.

Finally, instead of using \gls{ssl} directly to build surrogate models, it can also be employed to reduce the dimensionality of data fed to the surrogate model. Dang et al.~\cite{dang2024novel} use an \gls{ae} to derive representations of neural network architectures in the latent space. These lower-dimensional representations are then provided to a surrogate model for training purposes. After the surrogate model is trained, it is integrated into an evolutionary \gls{nas} system.

\section{Challenges and future research directions}
\label{sec:discussion_future_research}

\gls{essl} addresses challenges in representation learning and algorithmic optimisation. The intersection of \gls{ssl} and \gls{eml} can be beneficial to automate the design of neural networks when the labelled data is scarce~\cite{vinhas2024towards}, or when there is a need to design networks that can be transferred to multiple tasks~\cite{garcia2024progressive}. Additionally, \gls{essl} can be used to improve different aspects of \gls{ec} algorithms without depending on labels. Nevertheless, five main challenges were identified in this field, aiming to provide guidance on future work to be done within \gls{essl}.

\subsection{Pretext task design automation}

The most noticeable trend in \gls{essl} is the preference for auto-encoding models that rely goals based on input reconstruction, along with a lack of exploration into multiple branch networks. Although \gls{ae} models can be competitive, what these competitive proposals share is the incorporation of a masking objective instead of the original reconstruction-based loss~\cite{he2022masked, chen2024context}. This underscores the importance of researching the pretext task that produces the best possible representation. The works surveyed in this context approach pretext task learning as a tool to maintain model performance with more limited resources, optimise coefficients of a loss function associated with a single pretext task, or define a combinatorial optimisation problem by framing pretext task learning as a linear combination of multiple off-the-shelf pretext tasks. However, one possible avenue for future research could be exploring the evolution of new pretext tasks that could facilitate representation learning.

The design of pretext tasks, however, depends on the architecture chosen for the first stage. More specifically, \gls{ssl} architectures rely on an encoder component that is reused for the downstream task. Other components serve auxiliary roles only during the pretext stage, and their existence depends on the pretext task. In its entirety, the actual model architecture (encoder) is only a small part of a larger structure, which we will refer to as the meta-architecture from hereon. We propose that \gls{essl} could play a pivotal role in designing meta-architectures for pretext task solutions by providing a unified view. Siamese networks, multiple-branch networks, and \glspl{ae} can be seen as meta-architectures derived from a common definition but instantiated with different parameter sets. Therefore, it is possible to define a search space of meta-architectures that shapes how \gls{ec} algorithms explore candidate solutions. The design of pretext tasks under the meta-architecture space is naturally more intricate to define and search, but it has the potential of producing better representations. 

\subsection{Experiment design under label scarcity scenarios}

One of the main motivations for \gls{ssl} is the ability to learn representations when limited labelled data is available or when the labels are too costly to obtain. Our perception is that there is room for improvement in demonstrating this advantage through experimental design. Although some of the surveyed works include ablation studies to assess the impact of the available labelled data on the accuracy of a network, that is not done in a way that allows a fair comparison between approaches. Ultimately, \gls{essl} approaches can also be evaluated based on the amount of labelled data needed to obtain a minimum level of accuracy, meaning that one cannot evaluate the success of \gls{essl} works only based on the final accuracy. This requires an effort in defining a standardised methodology that evaluates the success of an \gls{essl} algorithm based on the dependency of labelled data.

When \gls{essl} is applied within the \gls{nas} field, evolved networks are generally subjected to post-evolution mechanisms to extract the best possible result. These mechanisms range between extending training of best individuals, retrain the best individual (or a skeleton networks built from the best individuals), or finetune representations. In such cases, when conducting ablation studies on the labelled data, it is important to keep consistency on the amount of labelled data between the evolution and post-evolution stages. If a limited percentage of labels is assumed, this assumption should be consistently applied throughout both the evolution (search phase) and post-evolution phases. For instance, conducting the search phase with a limited percentage of labels, followed by post-evolution improvements using the full labelled dataset, would likely produce competitive results. However, this does not support the argument of performance with scarce labels, and it adds unnecessary complexity compared to a purely Supervised Learning approach with the full labelled dataset, as noted by Rodrigues et al.~\cite{rodrigues2023performance}.

\subsection{Exploration of fitness metrics that balance the trade-off between time consumption and fidelity}

The output of an evolved Deep Neural Network evaluated on the pretext task (or on the downstream task if no fine-tuning occurs) is a set of features that can be reused across multiple downstream tasks. Therefore, one of the goals of \gls{essl} in the context of \gls{nas} is to search for a Deep Neural Network that produces representations applicable to multiple problems within a single evolutionary process. However, there is a trade-off between evaluating fitness on the downstream task versus the pretext task. When fitness is based on the downstream task, the metrics are high-fidelity, but the common evaluation bottlenecks reported in evolutionary \gls{nas} methods become more problematic, as a two-step learning process is introduced into the evolutionary cycle. By measuring fitness on the pretext task, we obtain lower-fidelity metrics depending on how similar the pretext and downstream tasks are. Moreover, the type of metrics one can use largely depends on the chosen pretext task. Pretext tasks like RotNet, which rely on explicit pseudo-labels, allow us to calculate accuracy on the pretext task and use it as fitness, which is theoretically more reliable than the validation loss from the pretext task. This is because relying on the loss function from the pretext task may not be a good proxy due to the randomness introduced during input sampling~\cite{li2021bossnas}. In contrast, \gls{ssl} methods like Barlow Twins, which reportedly produce better representations, do not provide the same pretext accuracy metric as RotNet, forcing us to either rely on the unreliable pretext validation loss or the time-consuming downstream accuracy. Therefore, further research is needed to develop methods that provide high-fidelity metrics that are less time-consuming, either through surrogate models or proxy metrics that do not require downstream task training~\cite{lu2023using, agrawal2022alpha}.

\subsection{Fitness evaluation speed-up}

When applying \gls{ssl} to \gls{ec}, the preference for \glspl{ae} with reconstruction loss is even more prominent. When \gls{ssl} targets evolutionary representation, \glspl{ae} are used to create a latent space that serves as the genotype. The goal in this case is to learn a mapping between genotype and phenotype, and vice versa. A major challenge here is that every time an individual needs to be evaluated, it must be decoded. One potential avenue for future research would be to develop reliable evaluation metrics directly from the representations, thereby eliminating the need for the decoding step and accelerating the evolutionary process. Regarding fitness assignment, the work of Wei et al.~\cite{wei2021self} and GenNAS~\cite{li2021generic, li2023extensible} provides valuable insights into how \gls{ssl} can generate surrogate models or proxies that reliably guide \gls{nas} algorithms while reducing reliance on labelled data. This line of research holds significant potential for future exploration. An ultimate goal would be to design a methodology that does not introduce substantial overhead and produces reliable, generic metrics that can be applied to any evolutionary \gls{nas} algorithm, regardless of the specific characteristics of the search space.

\subsection{Exploration of other meta-architectures}

As it was mentioned in earlier challenges, \glspl{ae} that are trained based on an input reconstruction objective are overall the most commonly adopted meta-architecture. When \gls{ssl} is applied for \gls{ec}, this trend is more clear. One exception to this rule is observed in the case when \gls{ssl} is applied to fitness evaluation aspects, as we did not find instances of \glspl{ae} being used to train surrogate models for evolutionary \gls{nas} fitness assignments. This is in contrast to the operators component, where no variation operators using meta-architectures beyond \glspl{ae} were observed, nor did we find any \gls{essl} works designing crossover operators. This highlights potential gaps in the current research that could be explored in future work.

Exploring other meta-architectures that might promote more robust representations is important to capture the true distribution of the data. In the the context of \gls{ec}, this has an impact in the mappings between the genotype and phenotype and in the variation operators, thereby impacting the success in finding the optimal solution. In the context of fitness assignment, searching for more robust representations can lead to fitness metrics with higher fidelity without impacting the time that it takes to evaluate a candidate solution.

\subsection{Summary}

The field of \gls{essl} has more research developed in cases whereby \gls{ec} is applied for \gls{ssl}. This is partially fuelled by the amount of studies that apply \gls{nas} approaches focused of self-supervised objectives. There is a need to explore alternative objectives and, more importantly, to engage \gls{ec} directly with the goal in the pretext task via loss function or the meta-architecture designed to train it. This highlights the challenge of using \gls{ec} to discover alternative forms of representation learning.

There is also a preference for \gls{ae} models that rely on an input reconstruction objective. From the perspective of applying \gls{ec} to \gls{ssl}, this raises the same aforementioned challenge. However, from the point view of applying \gls{ssl} for \gls{ec}, it also highlights that there can be alternative components designed with the help of \gls{ssl} models, which could improve aspects of \gls{ec}. 

Compared to supervised learning, using a model trained with \gls{ssl} comes with an overhead - it is a 2-step process whose pretext task can be time consuming to train. This means that if one relies on the downstream task to understand the quality of the representations, significant time costs will be incurred on any system that relies on this metric. Therefore, it is important to find reliable and computationally efficient proxy measures that can accurately reflect the quality of learned representations. 

When it comes to \gls{ssl}, the quality of any suggested approach should be measured against different amounts of labelled data, in order to prove its robustness to less labelled data. This consideration should be taken into account not only during the evolutionary process that comes up with a model, but also during any post-evolution methods that are applied to enhance the model obtained by the evolutionary system.

\section{Conclusions}
\label{sec:conclusion}

This survey provides a comprehensive review of the intersection between \gls{eml} and \gls{ssl}, introducing the emerging field of \glsreset{essl}\gls{essl}. We categorise the literature into two groups: (i) studies that apply \gls{ec} to \gls{ssl}, and (ii) studies that use \gls{ssl} to enhance \gls{ec}. Within each group, we further classify works according to the \gls{ssl} stage and the specific components of Deep Neural Networks or evolutionary algorithms targeted by \gls{ssl}.

Our analysis reveals two clear trends: first, most research applies \gls{ec} to \gls{ssl}, while the reverse direction remains underexplored; second, \glspl{ae} trained with input reconstruction objectives dominate current approaches. This indicates both their importance and the opportunity to investigate alternative or evolved \gls{ssl} objectives.


As this field develops, there is great potential for \gls{essl} to uncover novel techniques in representation learning and optimise evolutionary processes. By providing a structured overview of the field, highlighting its challenges, and identifying promising research directions, we aim to catalyse further exploration and development of \gls{essl} as a dynamic and interdisciplinary area of study.

\iftrue 

\section*{Acknowledgments}

This work is financed through national funds by FCT - Fundação para a Ciência e a Tecnologia, I.P., in the framework of the Project UIDB/00326/2025 and UIDP/00326/2025.

\fi

\bibliographystyle{elsarticle-num}
\bibliography{references.bib}

\begin{thebibliography}{100}
\expandafter\ifx\csname url\endcsname\relax
  \def\url#1{\texttt{#1}}\fi
\expandafter\ifx\csname urlprefix\endcsname\relax\def\urlprefix{URL }\fi
\expandafter\ifx\csname href\endcsname\relax
  \def\href#1#2{#2} \def\path#1{#1}\fi

\bibitem{banzhaf2023handbook}
W.~Banzhaf, P.~Machado, M.~Zhang, Handbook of Evolutionary Machine Learning, Springer, 2023.

\bibitem{zhang2011evolutionary}
J.~Zhang, Z.-h. Zhan, Y.~Lin, N.~Chen, Y.-j. Gong, J.-h. Zhong, H.~S. Chung, Y.~Li, Y.-h. Shi, Evolutionary computation meets machine learning: A survey, IEEE Computational Intelligence Magazine 6~(4) (2011) 68--75.

\bibitem{telikani2021evolutionary}
A.~Telikani, A.~Tahmassebi, W.~Banzhaf, A.~H. Gandomi, Evolutionary machine learning: A survey, ACM Computing Surveys (CSUR) 54~(8) (2021) 1--35.

\bibitem{mirjalili2019evolutionary}
S.~Mirjalili, H.~Faris, I.~Aljarah, Evolutionary machine learning techniques, Cham, Switzerland: Springer (2019).

\bibitem{kenton2019bert}
J.~D. M.-W.~C. Kenton, L.~K. Toutanova, Bert: Pre-training of deep bidirectional transformers for language understanding, in: Proceedings of naacL-HLT, Vol.~1, Minneapolis, Minnesota, 2019, p.~2.

\bibitem{tomasev2022pushing}
N.~Tomasev, I.~Bica, B.~McWilliams, L.~Buesing, R.~Pascanu, C.~Blundell, J.~Mitrovic, Pushing the limits of self-supervised resnets: Can we outperform supervised learning without labels on imagenet?, arXiv preprint arXiv:2201.05119 (2022).

\bibitem{chung2021w2v}
Y.-A. Chung, Y.~Zhang, W.~Han, C.-C. Chiu, J.~Qin, R.~Pang, Y.~Wu, W2v-bert: Combining contrastive learning and masked language modeling for self-supervised speech pre-training, in: 2021 IEEE Automatic Speech Recognition and Understanding Workshop (ASRU), IEEE, 2021, pp. 244--250.

\bibitem{wang2023videomae}
L.~Wang, B.~Huang, Z.~Zhao, Z.~Tong, Y.~He, Y.~Wang, Y.~Wang, Y.~Qiao, Videomae v2: Scaling video masked autoencoders with dual masking, in: Proceedings of the IEEE/CVF Conference on Computer Vision and Pattern Recognition, 2023, pp. 14549--14560.

\bibitem{hager2023best}
P.~Hager, M.~J. Menten, D.~Rueckert, Best of both worlds: Multimodal contrastive learning with tabular and imaging data, in: Proceedings of the IEEE/CVF Conference on Computer Vision and Pattern Recognition, 2023, pp. 23924--23935.

\bibitem{technologies9010002}
A.~Jaiswal, A.~R. Babu, M.~Z. Zadeh, D.~Banerjee, F.~Makedon, A survey on contrastive self-supervised learning, Technologies 9~(1) (2021).
\newblock \href {https://doi.org/10.3390/technologies9010002} {\path{doi:10.3390/technologies9010002}}.

\bibitem{gui2024survey}
J.~Gui, T.~Chen, J.~Zhang, Q.~Cao, Z.~Sun, H.~Luo, D.~Tao, A survey on self-supervised learning: Algorithms, applications, and future trends, IEEE Transactions on Pattern Analysis and Machine Intelligence (2024).

\bibitem{miikkulainen2023evolutionary}
R.~Miikkulainen, Evolutionary supervised machine learning, in: Handbook of Evolutionary Machine Learning, Springer, 2023, pp. 29--57.

\bibitem{wang2024application}
Y.~Wang, Q.~Zhang, G.-G. Wang, H.~Cheng, The application of evolutionary computation in generative adversarial networks (gans): a systematic literature survey, Artificial Intelligence Review 57~(7) (2024) 182.

\bibitem{de1994learning}
V.~R. de~Sa, Learning classification with unlabeled data, Advances in neural information processing systems (1994) 112--112.

\bibitem{bengio2013representation}
Y.~Bengio, A.~Courville, P.~Vincent, Representation learning: A review and new perspectives, IEEE transactions on pattern analysis and machine intelligence 35~(8) (2013) 1798--1828.

\bibitem{li2022understanding}
A.~C. Li, A.~A. Efros, D.~Pathak, Understanding collapse in non-contrastive siamese representation learning, in: European Conference on Computer Vision, Springer, 2022, pp. 490--505.

\bibitem{doersch2015unsupervised}
C.~Doersch, A.~Gupta, A.~A. Efros, Unsupervised visual representation learning by context prediction, in: Proceedings of the IEEE international conference on computer vision, 2015, pp. 1422--1430.

\bibitem{komodakis2018unsupervised}
N.~Komodakis, S.~Gidaris, Unsupervised representation learning by predicting image rotations, in: International conference on learning representations (ICLR), 2018.

\bibitem{larsson2016learning}
G.~Larsson, M.~Maire, G.~Shakhnarovich, Learning representations for automatic colorization, in: Computer Vision--ECCV 2016: 14th European Conference, Amsterdam, The Netherlands, October 11--14, 2016, Proceedings, Part IV 14, Springer, 2016, pp. 577--593.

\bibitem{7312476}
A.~Dosovitskiy, P.~Fischer, J.~T. Springenberg, M.~Riedmiller, T.~Brox, Discriminative unsupervised feature learning with exemplar convolutional neural networks, IEEE Transactions on Pattern Analysis and Machine Intelligence 38~(9) (2016) 1734--1747.
\newblock \href {https://doi.org/10.1109/TPAMI.2015.2496141} {\path{doi:10.1109/TPAMI.2015.2496141}}.

\bibitem{li2021generic}
Y.~Li, C.~Hao, P.~Li, J.~Xiong, D.~Chen, Generic neural architecture search via regression, Advances in Neural Information Processing Systems 34 (2021) 20476--20490.

\bibitem{zhang2021neural}
X.~Zhang, P.~Hou, X.~Zhang, J.~Sun, Neural architecture search with random labels, in: Proceedings of the IEEE/CVF conference on computer vision and pattern recognition, 2021, pp. 10907--10916.

\bibitem{pickup2014seeing}
L.~C. Pickup, Z.~Pan, D.~Wei, Y.~Shih, C.~Zhang, A.~Zisserman, B.~Scholkopf, W.~T. Freeman, Seeing the arrow of time, in: Proceedings of the IEEE Conference on Computer Vision and Pattern Recognition, 2014, pp. 2035--2042.

\bibitem{piergiovanni2020evolving}
A.~Piergiovanni, A.~Angelova, M.~S. Ryoo, Evolving losses for unsupervised video representation learning, in: Proceedings of the IEEE/CVF conference on computer vision and pattern recognition, 2020, pp. 133--142.

\bibitem{misra2016shuffle}
I.~Misra, C.~L. Zitnick, M.~Hebert, Shuffle and learn: unsupervised learning using temporal order verification, in: Computer Vision--ECCV 2016: 14th European Conference, Amsterdam, The Netherlands, October 11--14, 2016, Proceedings, Part I 14, Springer, 2016, pp. 527--544.

\bibitem{arandjelovic2017look}
R.~Arandjelovic, A.~Zisserman, Look, listen and learn, in: Proceedings of the IEEE international conference on computer vision, 2017, pp. 609--617.

\bibitem{korbar2018cooperative}
B.~Korbar, D.~Tran, L.~Torresani, Cooperative learning of audio and video models from self-supervised synchronization, Advances in Neural Information Processing Systems 31 (2018).

\bibitem{caron2018deep}
M.~Caron, P.~Bojanowski, A.~Joulin, M.~Douze, Deep clustering for unsupervised learning of visual features, in: Proceedings of the European conference on computer vision (ECCV), 2018, pp. 132--149.

\bibitem{yang2016joint}
J.~Yang, D.~Parikh, D.~Batra, Joint unsupervised learning of deep representations and image clusters, in: Proceedings of the IEEE conference on computer vision and pattern recognition, 2016, pp. 5147--5156.

\bibitem{zhuang2019local}
C.~Zhuang, A.~L. Zhai, D.~Yamins, Local aggregation for unsupervised learning of visual embeddings, in: Proceedings of the IEEE/CVF international conference on computer vision, 2019, pp. 6002--6012.

\bibitem{vincent2008extracting}
P.~Vincent, H.~Larochelle, Y.~Bengio, P.-A. Manzagol, Extracting and composing robust features with denoising autoencoders, in: Proceedings of the 25th international conference on Machine learning, 2008, pp. 1096--1103.

\bibitem{pathak2016context}
D.~Pathak, P.~Krahenbuhl, J.~Donahue, T.~Darrell, A.~A. Efros, Context encoders: Feature learning by inpainting, in: Proceedings of the IEEE conference on computer vision and pattern recognition, 2016, pp. 2536--2544.

\bibitem{zhang2017split}
R.~Zhang, P.~Isola, A.~A. Efros, Split-brain autoencoders: Unsupervised learning by cross-channel prediction, in: Proceedings of the IEEE conference on computer vision and pattern recognition, 2017, pp. 1058--1067.

\bibitem{he2022masked}
K.~He, X.~Chen, S.~Xie, Y.~Li, P.~Doll{\'a}r, R.~Girshick, Masked autoencoders are scalable vision learners, in: Proceedings of the IEEE/CVF conference on computer vision and pattern recognition, 2022, pp. 16000--16009.

\bibitem{srivastava2015unsupervised}
N.~Srivastava, E.~Mansimov, R.~Salakhudinov, Unsupervised learning of video representations using lstms, in: International conference on machine learning, PMLR, 2015, pp. 843--852.

\bibitem{chen2019self}
T.~Chen, X.~Zhai, M.~Ritter, M.~Lucic, N.~Houlsby, Self-supervised gans via auxiliary rotation loss, in: Proceedings of the IEEE/CVF conference on computer vision and pattern recognition, 2019, pp. 12154--12163.

\bibitem{noroozi2016unsupervised}
M.~Noroozi, P.~Favaro, Unsupervised learning of visual representations by solving jigsaw puzzles, in: Computer Vision--ECCV 2016: 14th European Conference, Amsterdam, The Netherlands, October 11-14, 2016, Proceedings, Part VI, Springer, 2016, pp. 69--84.

\bibitem{misra2020self}
I.~Misra, L.~v.~d. Maaten, Self-supervised learning of pretext-invariant representations, in: Proceedings of the IEEE/CVF conference on computer vision and pattern recognition, 2020, pp. 6707--6717.

\bibitem{chen2020simple}
T.~Chen, S.~Kornblith, M.~Norouzi, G.~Hinton, A simple framework for contrastive learning of visual representations, in: International conference on machine learning, PMLR, 2020, pp. 1597--1607.

\bibitem{dwibedi2021little}
D.~Dwibedi, Y.~Aytar, J.~Tompson, P.~Sermanet, A.~Zisserman, With a little help from my friends: Nearest-neighbor contrastive learning of visual representations, in: Proceedings of the IEEE/CVF International Conference on Computer Vision, 2021, pp. 9588--9597.

\bibitem{zbontar2021barlow}
J.~Zbontar, L.~Jing, I.~Misra, Y.~LeCun, S.~Deny, Barlow twins: Self-supervised learning via redundancy reduction, in: International Conference on Machine Learning, PMLR, 2021, pp. 12310--12320.

\bibitem{bardes2021vicreg}
A.~Bardes, J.~Ponce, Y.~LeCun, Vicreg: Variance-invariance-covariance regularization for self-supervised learning, arXiv preprint arXiv:2105.04906 (2021).

\bibitem{grill2020bootstrap}
J.-B. Grill, F.~Strub, F.~Altch{\'e}, C.~Tallec, P.~Richemond, E.~Buchatskaya, C.~Doersch, B.~Avila~Pires, Z.~Guo, M.~Gheshlaghi~Azar, et~al., Bootstrap your own latent-a new approach to self-supervised learning, Advances in neural information processing systems 33 (2020) 21271--21284.

\bibitem{chen2021exploring}
X.~Chen, K.~He, Exploring simple siamese representation learning, in: Proceedings of the IEEE/CVF conference on computer vision and pattern recognition, 2021, pp. 15750--15758.

\bibitem{zhou2021ibot}
J.~Zhou, C.~Wei, H.~Wang, W.~Shen, C.~Xie, A.~Yuille, T.~Kong, ibot: Image bert pre-training with online tokenizer, International Conference on Learning Representations (ICLR) (2022).

\bibitem{assran2023self}
M.~Assran, Q.~Duval, I.~Misra, P.~Bojanowski, P.~Vincent, M.~Rabbat, Y.~LeCun, N.~Ballas, Self-supervised learning from images with a joint-embedding predictive architecture, in: Proceedings of the IEEE/CVF Conference on Computer Vision and Pattern Recognition, 2023, pp. 15619--15629.

\bibitem{bardes2024revisiting}
A.~Bardes, Q.~Garrido, J.~Ponce, X.~Chen, M.~Rabbat, Y.~LeCun, M.~Assran, N.~Ballas, Revisiting feature prediction for learning visual representations from video, arXiv preprint arXiv:2404.08471 (2024).

\bibitem{correia2016xfaces}
J.~Correia, T.~Martins, P.~Martins, P.~Machado, X-faces: The exploit is out there, in: ICCC, 2016, pp. 164--171.

\bibitem{romera2024mathematical}
B.~Romera-Paredes, M.~Barekatain, A.~Novikov, M.~Balog, M.~P. Kumar, E.~Dupont, F.~J. Ruiz, J.~S. Ellenberg, P.~Wang, O.~Fawzi, et~al., Mathematical discoveries from program search with large language models, Nature 625~(7995) (2024) 468--475.

\bibitem{tian2020solving}
Y.~Tian, C.~Lu, X.~Zhang, K.~C. Tan, Y.~Jin, Solving large-scale multiobjective optimization problems with sparse optimal solutions via unsupervised neural networks, IEEE transactions on cybernetics 51~(6) (2020) 3115--3128.

\bibitem{garcia2024progressive}
C.~Garcia-Garcia, A.~Morales-Reyes, H.~J. Escalante, Progressive self-supervised multi-objective nas for image classification, in: International Conference on the Applications of Evolutionary Computation (Part of EvoStar), Springer, 2024, pp. 180--195.

\bibitem{vinhas2024towards}
A.~Vinhas, J.~Correia, P.~Machado, Towards evolution of deep neural networks through contrastive self-supervised learning, in: 2024 IEEE Congress on Evolutionary Computation (CEC), 2024, pp. 1--8.
\newblock \href {https://doi.org/10.1109/CEC60901.2024.10611910} {\path{doi:10.1109/CEC60901.2024.10611910}}.

\bibitem{barrett2023evolutionary}
N.~Barrett, Z.~Sadeghi, S.~Matwin, Evolutionary augmentation policy optimization for self-supervised learning, Adv. Artif. Intell. Mach. Learn. 3~(2) (2023) 1135--1164.
\newblock \href {https://doi.org/10.54364/AAIML.2023.1167} {\path{doi:10.54364/AAIML.2023.1167}}.

\bibitem{liu2024collaborative}
Y.~Liu, J.~Li, M.~Gong, H.~Liu, K.~Sheng, Y.~Zhang, Z.~Tang, Y.~Zhou, Collaborative self-supervised evolution for few-shot remote sensing scene classification, IEEE Transactions on Geoscience and Remote Sensing (2024).

\bibitem{li2022does}
X.~Li, J.~Shang, S.~Das, M.~Ryoo, Does self-supervised learning really improve reinforcement learning from pixels?, Advances in Neural Information Processing Systems 35 (2022) 30865--30881.

\bibitem{xue2021fast}
S.~Xue, H.~Chen, C.~Xie, B.~Zhang, X.~Gong, D.~Doermann, Fast and unsupervised neural architecture evolution for visual representation learning, IEEE Computational Intelligence Magazine 16~(3) (2021) 22--32.

\bibitem{jin2021automated}
W.~Jin, X.~Liu, X.~Zhao, Y.~Ma, N.~Shah, J.~Tang, Automated self-supervised learning for graphs, in: The Tenth International Conference on Learning Representations, {ICLR} 2022, Virtual Event, April 25-29, 2022, OpenReview.net, 2022.

\bibitem{han2022autonas}
Z.~Han, D.~Hong, L.~Gao, B.~Zhang, M.~Huang, J.~Chanussot, Autonas: Automatic neural architecture search for hyperspectral unmixing, IEEE Transactions on Geoscience and Remote Sensing 60 (2022) 1--14.

\bibitem{shen2024evolutionary}
X.~Shen, H.~Li, A.~Shankar, W.~Viriyasitavat, V.~Chamola, Evolutionary computation-based self-supervised learning for image processing: a big data-driven approach to feature extraction and fusion for multispectral object detection, Journal of Big Data 11~(1) (2024) 130.

\bibitem{li2023extensible}
Y.~Li, J.~Li, C.~Hao, P.~Li, J.~Xiong, D.~Chen, Extensible and efficient proxy for neural architecture search, in: Proceedings of the IEEE/CVF International Conference on Computer Vision, 2023, pp. 6199--6210.

\bibitem{rodrigues2023performance}
N.~Rodrigues, J.~Almeida, S.~Silva, Performance analysis of self-supervised strategies for standard genetic programming, in: Proceedings of the Companion Conference on Genetic and Evolutionary Computation, 2023, pp. 627--630.

\bibitem{david2014genetic}
O.~E. David, I.~Greental, Genetic algorithms for evolving deep neural networks, in: Proceedings of the companion publication of the 2014 annual conference on genetic and evolutionary computation, 2014, pp. 1451--1452.

\bibitem{lander2015evoae}
S.~Lander, Y.~Shang, Evoae--a new evolutionary method for training autoencoders for deep learning networks, in: 2015 IEEE 39th annual computer software and applications conference, Vol.~2, IEEE, 2015, pp. 790--795.

\bibitem{assuncao2018automatic}
F.~Assuncao, D.~Sereno, N.~Lourenco, P.~Machado, B.~Ribeiro, Automatic evolution of autoencoders for compressed representations, in: 2018 IEEE Congress on Evolutionary Computation (CEC), IEEE, 2018, pp. 1--8.

\bibitem{chen2020evolving}
X.~Chen, Y.~Sun, M.~Zhang, D.~Peng, Evolving deep convolutional variational autoencoders for image classification, IEEE Transactions on Evolutionary Computation 25~(5) (2020) 815--829.

\bibitem{sun2018evolving}
Y.~Sun, G.~G. Yen, Z.~Yi, Evolving unsupervised deep neural networks for learning meaningful representations, IEEE Transactions on Evolutionary Computation 23~(1) (2018) 89--103.

\bibitem{cai2018novel}
Y.~Cai, Z.~Cai, M.~Zeng, X.~Liu, J.~Wu, G.~Wang, A novel deep learning approach: Stacked evolutionary auto-encoder, in: 2018 International Joint Conference on Neural Networks (IJCNN), IEEE, 2018, pp. 1--8.

\bibitem{ludwig2023compressing}
O.~Ludwig, T.~Claes, Compressing wav2vec2 for embedded applications, in: 2023 IEEE 33rd International Workshop on Machine Learning for Signal Processing (MLSP), IEEE, 2023, pp. 1--6.

\bibitem{yan2024masked}
C.~Yan, X.~Chang, Z.~Li, L.~Yao, M.~Luo, Q.~Zheng, Masked distillation advances self-supervised transformer architecture search, in: The Twelfth International Conference on Learning Representations, 2024.

\bibitem{kanwal2021evolving}
S.~Kanwal, I.~Younas, M.~Bashir, Evolving convolutional autoencoders using multi-objective particle swarm optimization, Computers \& Electrical Engineering 91 (2021) 107108.

\bibitem{steinmetz2024st}
C.~J. Steinmetz, S.~Singh, M.~Comunit{\`{a}}, I.~Ibnyahya, S.~Yuan, E.~Benetos, J.~D. Reiss, {ST-ITO:} controlling audio effects for style transfer with inference-time optimization, in: Proceedings of the 25th International Society for Music Information Retrieval Conference, {ISMIR} 2024, San Francisco, California, {USA} and Online, November 10-14, 2024, 2024, pp. 661--668.
\newblock \href {https://doi.org/10.5281/ZENODO.14877423} {\path{doi:10.5281/ZENODO.14877423}}.

\bibitem{sors2021simple}
A.~Sors, R.~S. de~Rezende, S.~Ibrahimi, J.-M. Andreoli, Simple and effective balance of contrastive losses, arXiv preprint arXiv:2112.11743 (2021).

\bibitem{ramachandran2024clamp}
A.~Ramachandran, S.~Kundu, T.~Krishna, Clamp-vit: Contrastive data-free learning for adaptive post-training quantization of vits, in: European Conference on Computer Vision, Springer, 2024, pp. 307--325.

\bibitem{frumkin2023jumping}
N.~Frumkin, D.~Gope, D.~Marculescu, Jumping through local minima: Quantization in the loss landscape of vision transformers, in: Proceedings of the IEEE/CVF International Conference on Computer Vision, 2023, pp. 16978--16988.

\bibitem{ramachandran2024algorithm}
A.~Ramachandran, Z.~Wan, G.~Jeong, J.~Gustafson, T.~Krishna, Algorithm-hardware co-design of distribution-aware logarithmic-posit encodings for efficient dnn inference, in: Proceedings of the 61st ACM/IEEE Design Automation Conference, 2024, pp. 1--6.

\bibitem{hu2024joint}
C.~Hu, T.~Wu, C.~Liu, C.~Chang, Joint contrastive learning and belief rule base for named entity recognition in cybersecurity, Cybersecurity 7~(1) (2024) 19.

\bibitem{shang2024efficient}
R.~Shang, H.~Liu, W.~Li, W.~Zhang, T.~Ma, L.~Jiao, An efficient evolutionary architecture search for variational autoencoder with alternating optimization and adaptive crossover, Swarm and Evolutionary Computation 86 (2024) 101520.

\bibitem{cheng2020adaptive}
H.~Cheng, Z.~Wang, Z.~Wei, L.~Ma, X.~Liu, On adaptive learning framework for deep weighted sparse autoencoder: A multiobjective evolutionary algorithm, IEEE Transactions on Cybernetics 52~(5) (2020) 3221--3231.

\bibitem{sun2018particle}
Y.~Sun, B.~Xue, M.~Zhang, G.~G. Yen, A particle swarm optimization-based flexible convolutional autoencoder for image classification, IEEE transactions on neural networks and learning systems 30~(8) (2018) 2295--2309.

\bibitem{dimanov2021moncae}
D.~Dimanov, E.~Balaguer-Ballester, C.~Singleton, S.~Rostami, Moncae: Multi-objective neuroevolution of convolutional autoencoders, arXiv preprint arXiv:2106.11914 (2021).

\bibitem{hajewski2020evolving}
J.~Hajewski, S.~Oliveira, X.~Xing, Evolving deep autoencoders, in: Proceedings of the 2020 Genetic and Evolutionary Computation Conference Companion, 2020, pp. 123--124.

\bibitem{hajewski2021distributed}
J.~Hajewski, S.~Oliveira, X.~Xing, Distributed evolution of deep autoencoders, in: Intelligent Computing: Proceedings of the 2021 Computing Conference, Volume 1, Springer, 2021, pp. 133--153.

\bibitem{charte2019automating}
F.~Charte, A.~J. Rivera, F.~Mart{\'\i}nez, M.~J. del Jesus, Automating autoencoder architecture configuration: An evolutionary approach, in: International Work-Conference on the Interplay Between Natural and Artificial Computation, Springer, 2019, pp. 339--349.

\bibitem{charte2020evoaaa}
F.~Charte, A.~J. Rivera, F.~Mart{\'\i}nez, M.~J. del Jesus, Evoaaa: An evolutionary methodology for automated neural autoencoder architecture search, Integrated Computer-Aided Engineering 27~(3) (2020) 211--231.

\bibitem{wu2024evae}
Z.~Wu, L.~Cao, L.~Qi, evae: Evolutionary variational autoencoder, IEEE Transactions on Neural Networks and Learning Systems (2024).

\bibitem{sun2023multitask}
T.~Sun, Z.~He, Q.~Zhu, X.~Qiu, X.~Huang, Multitask pre-training of modular prompt for {C}hinese few-shot learning, in: A.~Rogers, J.~Boyd-Graber, N.~Okazaki (Eds.), Proceedings of the 61st Annual Meeting of the Association for Computational Linguistics (Volume 1: Long Papers), Association for Computational Linguistics, Toronto, Canada, 2023, pp. 11156--11172.
\newblock \href {https://doi.org/10.18653/v1/2023.acl-long.625} {\path{doi:10.18653/v1/2023.acl-long.625}}.

\bibitem{emm2024self}
T.~A. Emm, Y.~Zhang, Self-adaptive evolutionary info variational autoencoder, Computers 13~(8) (2024) 214.

\bibitem{tabak2024evolutionary}
G.~C. Tabak, D.~Molenaar, M.~Curi, An evolutionary neural architecture search for item response theory autoencoders, Behaviormetrika (2024) 1--24.

\bibitem{preen2021autoencoding}
R.~J. Preen, S.~W. Wilson, L.~Bull, Autoencoding with a classifier system, IEEE Transactions on Evolutionary Computation 25~(6) (2021) 1079--1090.

\bibitem{bu2024layer}
C.~Bu, Y.~Liu, M.~Huang, J.~Shao, S.~Ji, W.~Luo, X.~Wu, Layer-wise learning rate optimization for task-dependent fine-tuning of pre-trained models: An evolutionary approach, ACM Transactions on Evolutionary Learning 4~(4) (2024) 1--23.

\bibitem{hajewski2020evolutionary}
J.~Hajewski, S.~Oliveira, An evolutionary approach to variational autoencoders, in: 2020 10th Annual Computing and Communication Workshop and Conference (CCWC), IEEE, 2020, pp. 0071--0077.

\bibitem{hajewski2021efficient}
J.~Hajewski, S.~Oliveira, Efficient evolution of variational autoencoders, in: 2021 IEEE 11th Annual Computing and Communication Workshop and Conference (CCWC), IEEE, 2021, pp. 1541--1550.

\bibitem{rodriguez2019evolving}
L.~Rodriguez-Coayahuitl, A.~Morales-Reyes, H.~J. Escalante, Evolving autoencoding structures through genetic programming, Genetic Programming and Evolvable Machines 20 (2019) 413--440.

\bibitem{schofield2023genetic}
F.~Schofield, L.~Slyfield, A.~Lensen, A genetic programming encoder for increasing autoencoder interpretability, in: European Conference on Genetic Programming (Part of EvoStar), Springer, 2023, pp. 19--35.

\bibitem{ha2021evolving}
T.~Ha, X.~Gao, Evolving multi-view autoencoders for text classification, in: IEEE/WIC/ACM International Conference on Web Intelligence and Intelligent Agent Technology, 2021, pp. 270--276.

\bibitem{suganuma2018exploiting}
M.~Suganuma, M.~Ozay, T.~Okatani, Exploiting the potential of standard convolutional autoencoders for image restoration by evolutionary search, in: International Conference on Machine Learning, PMLR, 2018, pp. 4771--4780.

\bibitem{silhan2019evolution}
T.~Silhan, S.~Oehmcke, O.~Kramer, Evolution of stacked autoencoders, in: 2019 IEEE Congress on Evolutionary Computation (CEC), IEEE, 2019, pp. 823--830.

\bibitem{gustafson2017beating}
J.~L. Gustafson, I.~T. Yonemoto, Beating floating point at its own game: Posit arithmetic, Supercomputing frontiers and innovations 4~(2) (2017) 71--86.

\bibitem{chen2009local}
L.~Chen, A.~Buja, Local multidimensional scaling for nonlinear dimension reduction, graph drawing, and proximity analysis, Journal of the American Statistical Association 104~(485) (2009) 209--219.

\bibitem{zhou2024legal}
Y.~Zhou, X.~Yan, H.~Huang, H.~Yan, M.~Chen, Legal text retrieval with contrastive representation learning and evolutionary data augmentation, in: 2024 IEEE Congress on Evolutionary Computation (CEC), IEEE, 2024, pp. 1--7.

\bibitem{yang2006belief}
J.-B. Yang, J.~Liu, J.~Wang, H.-S. Sii, H.-W. Wang, Belief rule-base inference methodology using the evidential reasoning approach-rimer, IEEE Transactions on systems, Man, and Cybernetics-part A: Systems and Humans 36~(2) (2006) 266--285.

\bibitem{wei2021self}
C.~Wei, Y.~Tang, C.~N.~C. Niu, H.~Hu, Y.~Wang, J.~Liang, Self-supervised representation learning for evolutionary neural architecture search, IEEE Computational Intelligence Magazine 16~(3) (2021) 33--49.

\bibitem{freiberger2024fooling}
M.~Freiberger, P.~Kun, C.~Igel, A.~S. L{\o}vlie, S.~Risi, Fooling contrastive language-image pre-trained models with {CLIPM}asterprints, Transactions on Machine Learning Research (2024).

\bibitem{sacadura2024design}
R.~Sacadura, L.~Gon{\c{c}}alo, T.~Martins, P.~Machado, Expanding design horizons: Evolutionary tool for parametric design exploration with interactive and clip-based evaluation, in: M.~F. Santos, J.~Machado, P.~Novais, P.~Cortez, P.~M. Moreira (Eds.), Progress in Artificial Intelligence, Springer Nature Switzerland, Cham, 2025, pp. 78--90.

\bibitem{machado2024pixels}
P.~Machado, T.~Martins, J.~Correia, L.~Espírito~Santo, N.~Lourenço, J.~Cunha, S.~Rebelo, P.~Martins, J.~Bicker, From pixels to metal: Ai-empowered numismatic art, in: K.~Larson (Ed.), Proceedings of the Thirty-Third International Joint Conference on Artificial Intelligence, {IJCAI-24}, International Joint Conferences on Artificial Intelligence Organization, 2024, pp. 7717--7725, aI, Arts \& Creativity.
\newblock \href {https://doi.org/10.24963/ijcai.2024/854} {\path{doi:10.24963/ijcai.2024/854}}.

\bibitem{yu2023black}
L.~Yu, Q.~Chen, J.~Lin, L.~He, Black-box prompt tuning for vision-language model as a service., in: IJCAI, 2023, pp. 1686--1694.

\bibitem{xiao2024evolutionary}
J.~Xiao, K.~Yu, B.~Zhao, D.~Liu, Evolutionary neural architecture search with performance predictor based on hybrid encodings, in: 2024 14th International Conference on Information Science and Technology (ICIST), IEEE, 2024, pp. 854--859.

\bibitem{hu2024autoencoder}
Z.~Hu, Z.~Xiao, H.~Sun, H.~Yang, Autoencoder evolutionary algorithm for large-scale multi-objective optimization problem, International Journal of Machine Learning and Cybernetics (2024) 1--14.

\bibitem{wittenberg2020dae}
D.~Wittenberg, F.~Rothlauf, D.~Schweim, Dae-gp: denoising autoencoder lstm networks as probabilistic models in estimation of distribution genetic programming, in: Proceedings of the 2020 Genetic and Evolutionary Computation Conference, 2020, pp. 1037--1045.

\bibitem{wittenberg2023denoising}
D.~Wittenberg, F.~Rothlauf, C.~Gagn{\'e}, Denoising autoencoder genetic programming: strategies to control exploration and exploitation in search, Genetic Programming and Evolvable Machines 24~(2) (2023) 17.

\bibitem{wittenberg2022denoising}
D.~Wittenberg, F.~Rothlauf, Denoising autoencoder genetic programming for real-world symbolic regression, in: Proceedings of the Genetic and Evolutionary Computation Conference Companion, 2022, pp. 612--614.

\bibitem{reiter2023pretraining}
J.~Reiter, D.~Schweim, D.~Wittenberg, Pretraining reduces runtime in denoising autoencoder genetic programming by an order of magnitude, in: Proceedings of the Companion Conference on Genetic and Evolutionary Computation, 2023, pp. 2382--2385.

\bibitem{yuan2023particle}
G.~Yuan, B.~Wang, B.~Xue, M.~Zhang, Particle swarm optimization for efficiently evolving deep convolutional neural networks using an autoencoder-based encoding strategy, IEEE Transactions on Evolutionary Computation (2023).

\bibitem{cui2022bi}
M.~Cui, L.~Li, M.~Zhou, J.~Li, A.~Abusorrah, K.~Sedraoui, A bi-population cooperative optimization algorithm assisted by an autoencoder for medium-scale expensive problems, IEEE/CAA Journal of Automatica Sinica 9~(11) (2022) 1952--1966.

\bibitem{li2024enao}
Z.~Li, X.~Rao, S.~Liu, B.~Zhao, D.~Liu, Enao: Evolutionary neural architecture optimization in the approximate continuous latent space of a deep generative model, in: 2024 International Joint Conference on Neural Networks (IJCNN), IEEE, 2024, pp. 1--8.

\bibitem{gong2024bridge}
Y.~Gong, Y.~Sun, D.~Peng, X.~Chen, Bridge the gap between fixed-length and variable-length evolutionary neural architecture search algorithms, Electronic Research Archive 32~(1) (2024) 263--292.

\bibitem{shem2024deep}
E.~Shem-Tov, M.~Sipper, A.~Elyasaf, Deep learning-based operators for evolutionary algorithms, arXiv preprint arXiv:2407.10477 (2024).

\bibitem{churchill2014denoising}
A.~W. Churchill, S.~Sigtia, C.~Fernando, A denoising autoencoder that guides stochastic search, arXiv preprint arXiv:1404.1614 (2014).

\bibitem{probst2015denoising}
M.~Probst, Denoising autoencoders for fast combinatorial black box optimization, in: Proceedings of the Companion Publication of the 2015 Annual Conference on Genetic and Evolutionary Computation, 2015, pp. 1459--1460.

\bibitem{thakkar2019autoencoder}
S.~Thakkar, C.~Cao, L.~Wang, T.~J. Choi, J.~Togelius, Autoencoder and evolutionary algorithm for level generation in lode runner, in: 2019 IEEE Conference on Games (CoG), IEEE, 2019, pp. 1--4.

\bibitem{dang2024novel}
T.~Dang, T.~T. Nguyen, J.~McCall, K.~Han, A.~W.-C. Liew, A novel surrogate model for variable-length encoding and its application in optimising deep learning architecture, in: 2024 IEEE Congress on Evolutionary Computation (CEC), IEEE, 2024, pp. 1--8.

\bibitem{nguyen2015fooled}
A.~M. Nguyen, J.~Yosinski, J.~Clune, Deep neural networks are easily fooled: High confidence predictions for unrecognizable images, in: {IEEE} Conference on Computer Vision and Pattern Recognition, {CVPR} 2015, Boston, MA, USA, June 7-12, 2015, {IEEE} Computer Society, 2015, pp. 427--436.
\newblock \href {https://doi.org/10.1109/CVPR.2015.7298640} {\path{doi:10.1109/CVPR.2015.7298640}}.

\bibitem{chen2024context}
X.~Chen, M.~Ding, X.~Wang, Y.~Xin, S.~Mo, Y.~Wang, S.~Han, P.~Luo, G.~Zeng, J.~Wang, Context autoencoder for self-supervised representation learning, International Journal of Computer Vision 132~(1) (2024) 208--223.

\bibitem{li2021bossnas}
C.~Li, T.~Tang, G.~Wang, J.~Peng, B.~Wang, X.~Liang, X.~Chang, Bossnas: Exploring hybrid cnn-transformers with block-wisely self-supervised neural architecture search, in: Proceedings of the IEEE/CVF International Conference on Computer Vision, 2021, pp. 12281--12291.

\bibitem{lu2023using}
Y.~Lu, Z.~Liu, A.~Baratin, R.~Laroche, A.~Courville, A.~Sordoni, Using representation expressiveness and learnability to evaluate self-supervised learning methods, Transactions on Machine Learning Research (2023).

\bibitem{agrawal2022alpha}
K.~K. Agrawal, A.~K. Mondal, A.~Ghosh, B.~Richards, $\alpha$-req: Assessing representation quality in self-supervised learning by measuring eigenspectrum decay, Advances in Neural Information Processing Systems 35 (2022) 17626--17638.

\end{thebibliography}

\end{document}